\documentclass[14pt]{article}

\usepackage{PRIMEarxiv}
\usepackage{multirow}
\usepackage[utf8]{inputenc} 
\usepackage[T1]{fontenc}    
\usepackage{hyperref}       
\usepackage{url}            
\usepackage{booktabs}       
\usepackage{amsfonts}       
\usepackage{nicefrac}       
\usepackage{subcaption}
\usepackage{microtype}      
\usepackage{lipsum}
\usepackage{fancyhdr}       
\usepackage{graphicx}       
\graphicspath{{media/}}     
\usepackage{tabularx}
\usepackage{pdflscape}
\usepackage{algorithm}
\usepackage{algpseudocode}
\usepackage{amsmath}
\usepackage{amssymb}
\usepackage{color, xcolor}
\usepackage{soul}
\DeclareMathOperator*{\argmax}{arg\,max}

\title{Explanations-Driven Active Feature Acquisition for Algorithmic Recourse
}

\author{
  Vinura Galwaduge, Jagath Samarabandu \\
  Dept. of Electrical and Computer Engineering\\
  Western University \\
  London, ON, Canada\\
  \texttt{\{vgalwadu, jagath\}@uwo.ca} \\
}

\begin{document}
\maketitle

\begin{abstract}
Algorithmic recourse methods typically assume that a predictive model has access to all features of an individual. In practice, decisions are often made with partial information, because features are costly to acquire. Active feature acquisition addresses cost-constrained prediction, but existing methods are explanation-agnostic: prior work provides explanations only after acquiring additional features, rather than using explanations to drive acquisition. This work flips that and treats algorithmic recourse and feature acquisition jointly. We use Markov Blanket theory to unify counterfactual, semifactual, and alterfactual explanations and to characterize how available recourse grows as features are acquired. Building on this framework, we propose an Explanation-Driven Feature Acquisition (EDFA) method that selects features by explanatory value per unit cost. The framework is further extended with distribution-free validity guarantees for recourse issued from partial information, which signal trustworthy, lower-cost recourse, along with a lower bound on the calibration data required to certify them. Experiments on 7 publicly available datasets with neural network-based predictive models show that EDFA acquires substantially fewer features than state-of-the-art AFA baselines while maintaining comparable accuracy and yielding more decision-relevant, actionable recourse. The implementation is available on GitHub.
\end{abstract}


\section{Introduction}
Machine-learning models used as automated decision-making systems have become widespread across domains such as finance, healthcare, recidivism, and more. While these automated predictive models offer quick, first-hand predictions, they are often opaque about how they arrived at them~\cite{rudin_stop_2019}. Hence, explainability is often considered as an integral part of such models to elucidate the reasoning behind the automated decisions, and has been formalized in terms of their importance in certain regularization frameworks as well~\cite{gdpr}. Such explanations can clarify how and why the model made a given prediction, and what actions a user can take to alter or flip that prediction in the future. This type of feedback is generally known as `Algorithmic Recourse (AR)' and typically fall under the type of explanations called `Counterfactual Explanations (CF)'. Generally, CFs are considered as post-hoc explanations in the sense that they follow a prediction already made by a predictive model. Algorithmic recourse can be extended to other types of post hoc explanations related to CFs. These are termed as `Semifactual'(SF) and `Alterfactual'(AF) explanations. An SF is an explanation which does not flip the prediction, but further solidify it by providing similar, alternate instances where the prediction would still hold~\cite{keanyevenIf2023}. An AF explanation is similar to SF, but the it emphasizes the features that are not relevant to the prediction. As a summary, SFs and AFs do not provide actionable information to change a prediction; rather, they serve as complementary explanations to CFs.  The provision of explanations in the form of perturbations on a minimal set of features that cause the prediction by a model, is a common characteristic of all three aforementioned types of recourse. It is typically assumed that algorithmic recourse is provided for predictive models where all possible features are available to make a prediction (full information is available).\\

\begin{figure}[htbp]
    \centering
    \includegraphics[width=1\linewidth]{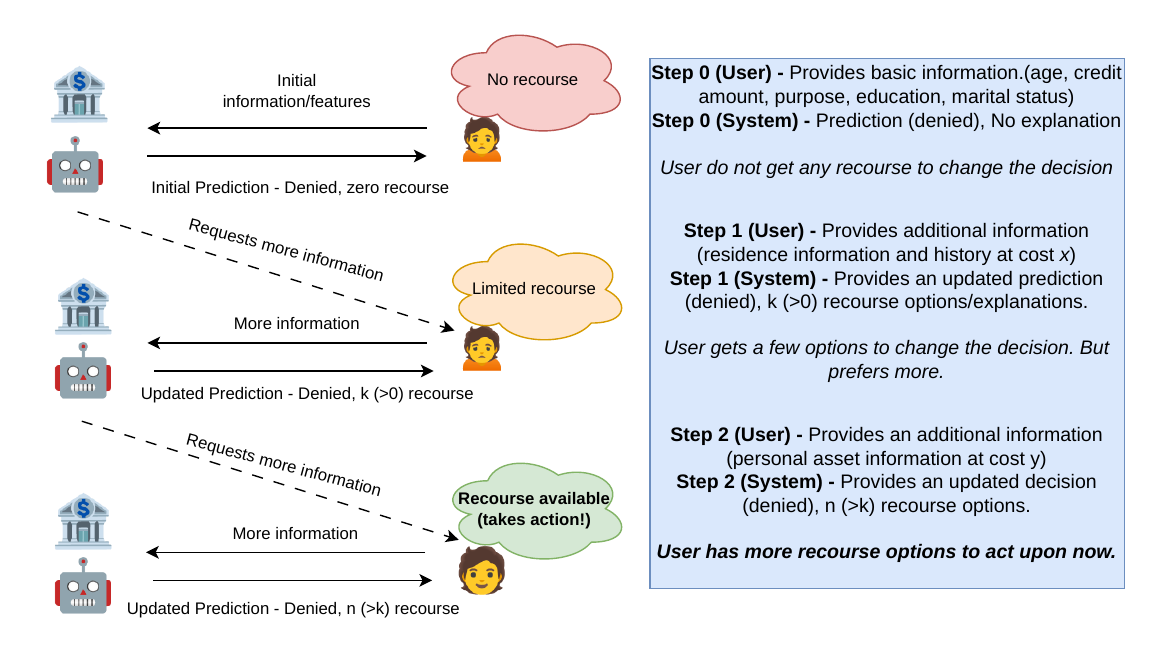}
    \caption{A credit application scenario showing how explanation drive feature acquisition helps.}
    \label{fig:fig1}
\end{figure}

However, in many real-world scenarios, predictions may have to be made with only a limited number of features (partial information) for a given instance. One main driver of such situations is the cost of feature acquisition. An example scenario is shown in figure~\ref{fig:fig1} where feature acquisition cost matters in the real world, where a user applies for credit (or a loan) at a financial institution. The user starts with only the required basic information and receives a denied verdict. However, the user does not provide all the information available to them initially because of privacy concerns, a lack of expertise to gather the required information, or time constraints. These limitations can be interpreted as subjective costs for the user in this scenario. The user then keeps providing more information at increasing costs, so that an actionable recourse can be arrived at.\\

\indent Active Feature Acquisition (AFA, also known as Dynamic Feature Acquisition) is a technique that addresses the challenge of acquiring features with required information that helps with the prediction, while accounting for the acquisition cost. AFA typically takes place during the inference stage of a predictive model. In particular, AFA is utilized in applications where high-dimensional data is prevalent and causes significant acquisition costs, such as medical diagnosis tasks~\cite{erion2022costaware}. In such scenarios, AFA aims to select features to achieve sufficient predictive performance for medical diagnosis under a given cost budget. It is known that not all the features would be required to provide a prediction to a given instance during inference time, and the required amount of features can differ from instance to instance as well. Beyond the requirement of a prediction, explanations also drive the number of features required. However, conventional AFA disregards the requirement of explanations. In scenarios such as the ones shown in figure~\ref{fig:fig1}, AFA would govern the number of features picked to make a prediction. This inadvertently affects explanation quality because explanations rely on the available, acquired features. In a credit application scenario such as in figure~\ref{fig:fig1}, the user may not opt to provide all the information at the beginning and perceive certain costs of providing information.
Based on this observation, exploring the interconnectedness between AFA and recourse is the main goal of our work. In particular, we aim to design a framework that treats explainability as an additional requirement of the AFA process and to determine whether imposing this constraint yields better explanations.



Under this broad theme, we conduct extensive experiments on 7 real-world tabular datasets from financial and medical domains. A novel, explanation-driven active feature acquisition method is introduced, and 3 AFA methods from the literature are tested as baseline AFA methods. To the best of our knowledge, this is the first work that treats explanations (algorithmic recourse) as an objective of AFA, rather than merely explaining AFA decisions. The following contributions are made in this paper.   

\begin{enumerate}
    \item We propose a novel, Explanation-Driven Active Feature Acquisition method (EDFA) using a Markov Blanket (MB) theory-based framework. We show that the same framework can unify three types of recourse; Counterfactual (CF), Semi-factual (SF) and Alterfactual (AF) Explanations.

    \item We show that this framework is able to utilize semifactuals which were previously treated as passive complements to counterfactuals in the literature, to provide signals identifying which unacquired features unlock new recourse. As shown in Proposition 1, EDFA's acquisition policy operationalizes this semifactual signaling principle.
    
    \item We propose a technique to find the minimal, sufficient dataset size that is needed for proper calibration and the calculation of validation guarantees for early recourse (recourse derived from limited number of features) even when all the features are available. These validation guarantees signal whether to act on early recourse, given limited available information, rather than waiting for all features to be available. 
    
    \item The proposed EDFA method and the validation guarantees are theoretically proven, and empirically verified using state-of-the-art methods, on real-world datasets. 
\end{enumerate}

The extensive set of experiments shows that the proposed EDFA method produces explanations (or recourse) that are lower cost, more decision-relevant and actionable than the baseline AFA methods, while maintaining comparable predictive performance to that of the baseline methods. Furthermore, the validation guarantees act as signals that identify lower-cost recourse options which are easier to act upon. Furthermore, EDFA prevents possible manipulations of the predictive models through explanations, where certain users may find features through explanations, that help them to game the predictions. As EDFA picks more decision-relevant features than the baseline AFA methods, the room for such manipulations can be reduced.\\ 
The rest of the paper is structured as follows. In Section~\ref {sec:theory}, the theoretical background, including the derivations of our proposed method, is provided. Section~\ref{sec:lit} covers the related prior work. Section~\ref{sec:method} provides the details of our proposed method and the experimental setup. Results and in-depth discussion are in sections~\ref{sec:results} and~\ref{sec:discussion}. Section~\ref{sec:conclusion} contains the concluding remarks. 
   

\section{Related Work}
\label{sec:lit}
\subsection{Markov Blanket (MB) Discovery}
MB is regarded as the optimal predictive feature set in a supervised learning setting~\cite{yaramakalaSpeculative2005, brown2012}. Incremental Association Markov Blanket (IAMB) is a classic MB discovery algorithm that greedily discovers the MB based on conditional independence tests~\cite{Tsamardinos2003AlgorithmsFL}. More data efficient MB discovery algorithms have been proposed later, which do not require large conditioning sets to do reliable MB discovery~\cite{lingBAMB2019}. It has been shown that the MB can be used to recover a causal, Partial DAG (PDAG) itself, using additional steps to remove extra edges that are not direct causal links in the discovered MB (i.e., edges that are direct spouse links to the target variable; $Sp(Y)-Y$)~\cite{pelletUsing2008}. Three MB discovery algorithms are used in this work as distinct approaches to MB search. They are described in more detail in the methodology section~\ref{sec:method}\\

\subsection{Active Feature Acquisition (AFA)}
AFA is a well-known paradigm for sequentially acquiring features under a given budget constraint to make accurate predictions. AFA is also termed as `Dynamic Feature Acquisition', which distinguishes it from static feature selection methods~\cite{guyonetal2003}. While static feature selection methods consistently pick a specific subset of features for all instances, AFA dynamically selects features for a given instance under a budget constraint while accounting for already acquired features. The link between information-theoretic feature selection and MB theory was formally introduced over a decade ago, where it is shown that MB search algorithms such as IAMB~\cite{tsamardinosPrincipled2003} correspond to feature selection by sequentially searching by maximizing conditional likelihood (i.e.- adding features that have the maximum mutual information with the current selected features)~\cite{brown2012}. AFA methods can be broadly categorized into two main types; Markov Decision-based AFA methods and greedy approaches. 
The literature has already acknowledged the requirement to study the link between local explanation methods and AFA~\cite{aronssonSurvey2026}. However, most work has addressed only one side of this requirement: how to explain the acquisitions made by an AFA algorithm. `Explainable AFA' is the specific branch of AFA that caters to this requirement. The concept of sub-goals has been used to explain acquisitions, where each acquisition step can be understood as completing a well-defined subgoal~\cite{liDistribution2024a}. In other work, the acquisition policy itself is considered as interpretable by design~\cite{chattopadhyayInterpretable2023, chattopadhyayVariational2023}. More recent AFA algorithms, such as Discriminative Mutual Information Estimation (DIME), perform comprehensive evaluations of AFA in scenarios including variable cost assignments per feature and per instance~\cite{gadgil2024estimating}.

\subsection{Algorithmic Recourse}
The concept and methods of algorithmic recourse were introduced as a means of obtaining actionable explanations for decisions made by automated predictive models~\cite{wachter_counterfactual_2018}. The requirement of algorithmic recourse was further emphasized by various regulatory frameworks, notably the `General Data Protection Act' (GDPR)~\cite{korikovCounterfactualExplanationsOptimizationBased2021}. A multitude of different counterfactual explanation (CF) methods that generate algorithmic recourse have been proposed for many application domains and contexts~\cite{verma_counterfactual_nodate, karimi_survey_2023, guidotti_counterfactual_2022, zhangSurveyCounterfactualExplanations2023}. Many aspects pertaining the efficacy of algorithmic recourse have been discussed, such as robustness~\cite{sharmaCERTIFAICommonFramework2020, dutta_robust_2022, virgolin_robustness_2023}, privacy~\cite{pawelczyk_privacy_2022, pentyala_privacy-preserving_2023}, fairness~\cite{schoffer_interplay_2023}, causality~\cite{KarKugSchVal20} and how they interact with each other~\cite{venkatasubramanian_philosophical_2020, barocasHiddenAssumptionsCounterfactual2020}. A handful of proposals aim to account for many real-world manifestations of the aforementioned aspects. One such relevant scenario to this work is called sequential recourse, where users would follow a gradual implementation of the actions suggested by the CFs~\cite{poyiadziFACEFeasibleActionable2020, small_counterfactual_2023, wangReinforcedPathReasoning2024, yamao_distribution-aligned_2025}. There, the user is expected to reach their preferred goal through a series of actions suggested by (possibly multiple) recourse options rather than achieving it once. This sequential decision process involves nuances, where a user may fail to fully implement a suggested action and requires new, additional recourse in order to reach the goal, which is termed as `iterative partial fulfillment'~\cite{zhouIterativePartialFulfillment2023a}. Another scenario is when some user information is missing due to data collection errors or privacy concerns, where conventional imputation methods have been shown to be inadequate~\cite{kanamori_algorithmic_2024}. Diversity of CFs is another crucial factor that enables users to choose among different plans of action and thus exercise autonomy in the process~\cite{sokol_one_2020}. Users may also have preferences over certain actions and perceived action costs, which requires recourse tuned to those preferences~\cite{buddeUniversal2026}.\\
Acquiring knowledge of causal relationships among features is considered important for algorithmic recourse (e.g., through Structured Causal Models). However, recovering it is not always feasible~\cite{KarKugSchVal20}. Hence, other proxy measures have also been used, such as plausibility~\cite{guidotti_counterfactual_2022,delserGeneratingTrustworthyCounterfactual2024, smyth2022few}, and uncertainty measurements~\cite{sokolAll2025}.\\ 
While CFs are the prominent type of explanation used for recourse, semi-factual explanations (SF)~\cite{aryalEvenIfExplanations2023, Alfano_Greco_Mandaglio_Parisi_Shahbazian_Trubitsyna_2025} and alter-factual explanations (AF)~\cite{nguyen2025nomatterxai} are proposed as complementary explanation types to CFs. Out of the two, SFs have been more widely discussed and have been proposed as a type of explanation that further strengthens the decision made by a model. In particular, SFs have been proposed as a mechanism to solidify positive outcomes obtained from a predictive model~\cite{keanyevenIf2023}. However, this strengthening effect applies to negative outcomes as well, preventing users from finding recourse. As for the AFs, they are utilized as explanations that emphasize the irrelevant features of a decision~\cite{mertes2024}.\\

Based on the prior work, the closest approaches to our current work are sequential recourse and recourse under missing information. However, the former assumes that predictive models already have access to all information about a user, whereas the latter focuses on the more specialized case where certain features are not available to the predictive model at all, and it must provide recourse options with only limited information. The idea of personalized recourse is also related to our proposed framework, which is based on the fact that the constraints of the CF generation problem is subject to individual preferences, but not the feature-level information available to the predictive model. However, our proposed frameworks helps develop a system where individual preferences or perceived cost of features are accounted for.

\section{Theoretical Background}
\label{sec:theory}

\subsection{Markov Blanket Theory}

The Markov Blanket (MB) theory defines a set of random variables (or features; depicted by nodes) that isolates a given random variable from the other random variables outside that set.

\textbf{Definition 1 - Markov Blanket: }Let $G=(V,E)$ be a directed acyclic graph (DAG) over a set of random variables $V=\{Y,X_1,X_2,...,X_n\}$ with a joint distribution of $P$. The Markov Blanket of $Y$ is $MB(Y)=Pa(Y) \bigcup Ch(Y) \bigcup Sp(Y)$, where $Sp(Y) = \bigcup_{C \in Ch(Y)} Pa(C) \backslash {Y}$.\\

\textbf{Definition 2 - d-separation:}
Two sets of variables $X_1$ and $X_2$ are \textit{d-separated} by another set $X_3$ in DAG $G$, if there exist chain ($X_i \rightarrow X_m \rightarrow X_j$), a fort ($X_i \leftarrow X_m \rightarrow X_j$) where $X_m \in X_3$ or a collider ($X_i \rightarrow X_m \leftarrow X_j$) where $X_m \notin X_3$.

\textbf{Assumption 1 - Causal Markov Condition:}  Every variable in $V$ is conditionally independent of its non-descendants, given its parents ($Pa$) is $G$.

\textbf{Assumption 2 - Faithfulness:}  Every conditional independence in $P$ is entailed by \textit{d-separation} in $G$.

\begin{figure}[htbp]
    \centering
    \includegraphics[width=0.5\linewidth]{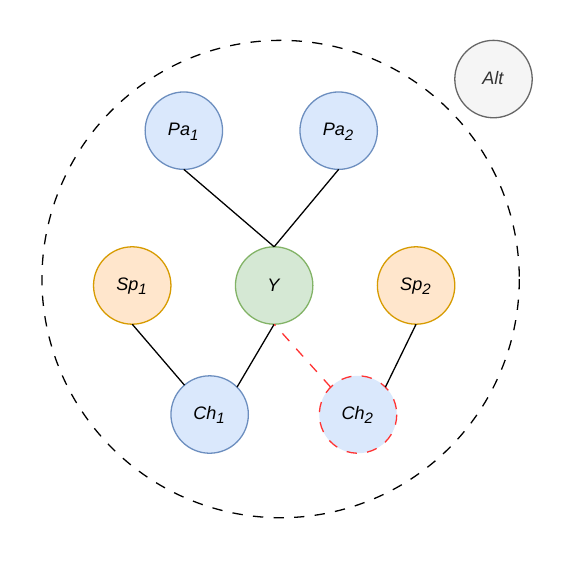}
    \caption{A sample MB with 2 parents ($Pa$), 2 spouses ($Sp$), and 2 child nodes ($Ch$) of $Y$. The dotted circle depicts the MB which separates the MB from the non-MB ($alt$) nodes. The solid edges depict the features that are currently observed. The child node $Ch_2$ has not been observed yet.}
    \label{fig:figMB}
\end{figure}

\textbf{Assumption 3 - Aligned model:} 
The first two assumptions are the default assumptions for MB theory. The 3rd assumption is that an ideal model would be trained on MB features. To better explain these concepts, figure~\ref{fig:figMB} shows a simple MB, including a \textit{d-separation} scenario.

\subsection{Explainability and Markov Blanket theory}

To study how algorithmic recourse is theoretically connected to active feature acquisition, we derive theoretical results from MB theory. The definitions below are provided as the foundations of our theoretical results.

\textbf{Definition 3 - Actual Cause:} ~\cite{pearl2016} A subset of features is an actual cause of a model prediction ($f(x)$), given a 

\textbf{Definition 4 - MB unit:} An MB unit of $Y$ is a subset $U = \{PC_j\} \bigcup Sp(PC_j) \subset MB(Y)$, where $PC_j \in Pa(Y) \bigcup Ch(Y)$ and $Sp(PC_j)$ are the spouse nodes linked through $PC_j$. \\

\textbf{Definition 5 - Observational Necessity:}  A feature subset $S$ is considered observationally necessary~\cite{kommiyamothilalUnifyingFeatureAttribution2021} for a prediction $y^*$ at instance $x$ if they flip the prediction when perturbed ($x'$) while keeping the other feature values unperturbed ($x_{\bar S}$):
\begin{equation}
\alpha(S, x, y^*)
  = \mathbb{E}_{x'_S,\, x_{\bar S}}
    \!\left[\, \mathbb{1}\!\left[\, f(x'_S, x_{\bar S}) \neq y^* \,\right] \right] >0
\end{equation}\\

\textbf{Definition 6 - Observational Sufficiency :} 
A feature subset $S$ is considered observationally sufficient~\cite{kommiyamothilalUnifyingFeatureAttribution2021} for a prediction $y^*$ at instance $x$ if they yield the prediction with any value for the other features ($x_{\bar S}$)::

\begin{equation}
\beta(S, x, y^*)
  = \mathbb{E}_{x'_S,\, x_{\bar S}}
    \!\left[\, \mathbb{1}\!\left[\, f(x_S, x'_{\bar S}) = y^* \,\right] \right]=1
\end{equation}

\textbf{Definition 7 - Tripartite Recourse:}  Given an acquired feature set $A \subseteq V \backslash{\{Y\}}$, observed values of the features $x_A$, and the prediction $y^*=f(x_A)$, the tripartite recourse $\mathcal{E}(A,x_A,y^*)$ partitions $V\backslash{\{Y\}}$ into three sets:

\textbf{Counterfactual Set :}
Counterfactual explanations emphasizes the feature changes that would flip or change the original prediction made by the decision making model on a given data instance.\\
As per the definition of observational necessity, these are the currently observed MB features that would flip the prediction if perturbed.

\begin{equation}
    \mathcal{C}(A) = \{X_i \in MB(Y) \cap A: \alpha(\{X_i\},x,y^*) >0\}
\end{equation}

\textbf{Semifactual Set :}
Semifactual explanations emphasize the feature changes that would still result in the original prediction made by the decision-making model on a given data instance.\\
The semifactual feature set ($\mathcal{S}(A)$) is the set of MB features that are currently not necessary. These features are currently unobserved or have a blocked causal pathway to $Y$ (e.g., a spouse node whose shared child is unobserved).

\begin{equation}
    \mathcal{S}(A) = \{X_i \in MB(Y) : X_i \notin A~or~\alpha(\{X_i\},x,y^*)\backslash{\mathcal{C(A)}} =0\}
\end{equation}

\textbf{Alterfactual Set :}
Alterfactual explanations emphasize the feature changes that are not relevant to the original prediction made by the decision-making model on a given data instance.\\
These are the non-MB features that are irrelevant for the decision made by the model at any stage, regardless of what is observed.

\begin{equation}
    \mathcal{A}_{\text{alt}}= V\backslash{\{Y\}}\backslash{MB(Y)}
\end{equation}

\textbf{Definition 8 - Counterfactual Recourse Frontier:} The recourse frontier at an acquisition state $A_i$ quantifies the number of features that produces valid counterfactuals.

\begin{equation}
    \mathcal{R}(A_i) = \{(X_i,x'_i):X_i \in \mathcal{C}(A_i),\}
\end{equation}

\textbf{Definition 9 - Latent Recourse:} A semifactual feature $X_i \in \mathcal{S}(A)$ has latent recourse if there exists unobserved features $X_j \notin A$ such that acquiring it reveals new counterfactual recourse pathways by transitioning $X_i$ from $\mathcal{S}$ to $\mathcal{C}$.
\begin{equation}
    \mathbb{E}\left[\, \left[|\mathcal{R}(A \bigcup \{X_j\})|\right] \right] > |\mathcal{R(A)}|
\end{equation}

\textbf{Theorem 1 - MB Sufficiency and Necessity:} Under assumptions 1 and 2, MB provides features that fulfill necessity and sufficiency conditions;
\begin{enumerate}
    \item MB(Y) is observationally sufficient: $\beta (MB(Y),x,y^*)=1, \forall x$
    \item Every $X_i \in MB(Y)$ is \emph{probabilistically necessary} as it remains statistically dependent on $Y$ given the remaining blanket features, $I(Y;X_i \mid X_{MB(Y) \backslash \{X_i\}})>0$. If, in addition, $X_i$ is
  label-flipping over its probabilistically plausible feature values (label flipping condition, appendix~\ref{sec:proof}), then
  there exists an instance $x$ with feature values such that $\alpha(\{X_i\},x,y^{*})>0$. Hence, they are observationally necessary (definition 5, Observational Necessity) as well.
    \item $\forall X_i \in \mathcal{A}_\text{alt}: \alpha(\{X_i\},x,y^*)=0, \forall x$
\end{enumerate}

It should be noted that this means that although the MB contains probabilistically necessary features, not all of them can be label flipping (or counterfactually active) features. An MB feature whose plausible feature values that never crosses the decision boundary has $\alpha=0$ and is a semifactual feature.

\textbf{Proposition 1 - Semifactual to Counterfactual transition:} Let $U=\{Ch_j \bigcup Sp(Ch_j)$ be an MB unit where $Ch_j \in Ch(Y)$ is a child, and let $Sp_k \in Pa(Ch_j) \backslash{Y}$ be a spouse of $Y$ under assumptions 1 and 2.

\begin{enumerate}
    \item Before observing $Ch_j$: If $Ch_j \notin A$ and no descendant of $Ch_j$ is in $A$, then $I(Y;X_{Sp_k}|X_A=x_A)=0, $ and $Sp_k \in S(A)$ (a semifactual feature)
    \item After observing $Ch_j$: If $Ch_j \in A$, then $I(Y;X_{Sp_k}|X_A=x_A)>0$, and $Sp_k$ may transition to $\mathcal{C}(A)$ at instances.
\end{enumerate}

\textbf{Remark 1 - Semifactuals as Acquisition Signals:} Proposition 1 establishes that semifactual features are not only currently irrelevant features, but also provide information for acquisition signals. A semifactual spouse $Sp_k$ implicitly points to the unobserved $Ch_j$ whose acquisition would activate the explaining away effect and potentially reveal new counterfactual recourse options.

\textbf{Proposition 2 - Parent, Child nodes (PC)-Priority  under Conditional Mutual Information (CMI)
based Acquisition}
Under assumptions 1 and 2, if the acquired set $A$ does not contain the shared child $Ch_j$, then:
\begin{equation}
    I(Y; X_{Ch_j}|X_A=x_A) >0 = I(Y;X_{Sp_k}| X_A = x_A)
\end{equation}
A CMI-based policy (a policy that will always prioritize CMI gain) acquires the PC member before the spouse within an MB unit in such settings. Acquiring $PC$ first weakly enlarges the one-step recourse frontier and strictly enlarges it whenever the label-flipping condition holds for some unlocked $Sp$ (Proposition 1). PC-first therefore, weakly dominates spouse-first for recourse-frontier growth within a unit.

Proofs of Theorem 1, Proposition 1 and 2 are provided in the appendix (section~\ref{sec:proof}).

\subsection{Probabilistic Bounds of Recourse Validity}

This section primarily discusses the theoretical foundations for deriving the validity guarantees for the recourse provided by the EDFA method.\\
Ideally, a validity guarantee for a recourse obtained for a given test instance would provide a probabilistic, confidence (upper) bound under which the validity of the said recourse will be preserved. However, it is known that providing such instance-specific guarantees is not possible, and often the bounds are distribution-specific~\cite{foygelbarberLimits2021}. On the other hand, techniques such as Conformal predictions only provide a bound based on the expected validation rate calculated over a calibration set~\cite{angelo2023}. Hence, Risk Controlling Prediction Sets (RCPS) are utilized in our proposed framework~\ref{eq:bound0}~\cite{bates2021}.

\begin{equation}
\label{eq:bound0}
    P(R(\hat \tau) \leq \alpha) \geq 1-\delta
\end{equation}
Here, the probabilistic guarantee stands for the observed (or measured) risk being lower than a certain risk level($\alpha$) with more than $1-\delta$ probability. The term $\delta$ stands for an error rate that this guarantee may not hold. The threshold $\hat \tau$, which is the variable that is monitored, is calculated as follows,
\begin{equation}
\label{eq:bound1}
    \hat \tau = max\{\tau \in T: UCB(\hat R(\tau);n. \delta) \leq \alpha\}
\end{equation}  

\begin{equation}
\label{eq:hb}
    P(R,n,\mu) = min(e\mathbb{P}[Bin(n,\mu) \leq \lceil nR \rceil], e^{-2n(\mu-R)^2})
\end{equation} 
$UCB$ (in eq.~\ref{eq:bound1}) stands for the `Upper Confidence Bound'. `Hoeffding-Bentkus (HB)' UCB variant is used in our proposed framework. Hoeffding theory provides a natural expression for an `upper bound of the sum of random variables deviating from its expected value by more than a certain amount (risk)'~\cite{Hoeffding01031963}. `Bentkus' inequality theory proposes an improvement on Hoeffding~\cite{Benktus2004}. We do not aim to provide a detailed description of the HB inequalities in this work. HB terms are shown in eq.~\ref{eq:hb}, where the two terms are the `Bentkus' and `Hoeffding' terms, respectively.

The $\tau$ term denotes the threshold derived from the prediction confidence of the model. $\hat \tau$ stands for the confidence threshold, under which RCPS-HB certifies the mean loss ($\hat{R}(\tau)$) to be less than the specified $\alpha$ invalidation level, with an error rate of $\delta$. In other words, the RCPS-HB returns a prediction confidence-based threshold, under 
Per instance validity is measured using the below loss function.

\begin{equation}
\label{eq:prob1}
    L_i(\tau) = \frac{1}{R}\sum_{r=1}^{R}
    \Bigl[\, \hat y_{i,k} \neq \hat y_{i,K_{i,r}}
    \ \lor \ CF_{k_r}\ \text{invalid}@K_{i,r} \,\Bigr]
\end{equation}
In the two terms joined by the OR operator, the first term quantifies whether the prediction made at step $k$ stays consistent at the last step ($k=K$). The second term quantifies whether the recourse obtained at step $k$ is valid at step $k=K$. The first term is crucial for checking whether the prediction itself remains consistent, which otherwise adds an additional layer of confusion for the user about whether to trust it. Then, we calculate the empirical risk at a given threshold as follows, where $n$ is the number of data samples in the calibration set.

\begin{equation}
    \hat{R}(\tau) = \frac{1}{n}\sum_{i=1}^{n} L_i(\tau)
\end{equation}

The population risk can then be approximated as,$\mathbb{E}(\hat{R}(\tau)))$. $\hat{\tau}$ is the maximum (corresponds to earlier steps of AFA) threshold that is empirically found, such that the validation guarantee is also preserved (eq.~\ref{eq:bound1}). The lowest possible risk is denoted by $\hat{R}[0]$.\\
The number of calibration data points ($n$) required to derive the UCB is as follows. From the Hoeffding term in equation~\ref{eq:hb},

\begin{equation}
    UCB_H(R,n,\delta) = \hat{R}+\sqrt{\frac{ln(1/\delta)}{2n}} \leq \alpha
\end{equation}

solving for $n$ gives a lower bound for the required for $n$, as below. However, this is a stricter bound since the Binomial Bentkus term is left out (from eq~\ref{eq:hb}). For this reason, we adopt the empirically calculated term, as shown in the results below.

\begin{equation}
\label{eq:hb-lower}
    n_{required} \geq \frac{ln(1/\delta)}{2(\alpha-\hat{R}[0])^2}
\end{equation}

\section{Methodology}
\label{sec:method}
\subsection{Explanation Driven active Feature Acquisition (EDFA)}
In addition to the assumptions mentioned in section~\ref{sec:theory}, we make the following additional assumptions to support the proposed method (Explanation Driven Feature Acquisition - EDFA).\\

\textbf{Assumption 4 - Known MB with MB Unit Structure:} MB(Y) and the PC-Spouse
mapping dictionary $A = {PC_j \mapsto Sp(PC_j)}^k_{j=1}$ are identified from training data prior to test-time feature acquisition.

\textbf{Assumption 5 - Feature Actionability:} The members in the counterfactual set ($C(A)$) certifies that perturbing them flips the model prediction. For simplicity, the features acquired and within the MB structure are considered actionable in this work. Sensitive features are always excluded from the actionable features.

Algorithm~\ref{alg:edfa} shows the proposed EDFA method. The core idea behind EDFA is that each feature acquisition step is based on both the cost and the feature's explanatory power. As shown in the algorithm, acquisition decisions depend on whether MB-units are partially acquired or not yet started (no node in the MB-unit has been acquired yet). Two routes turn semifactual features into counterfactual features (Definition 9). One route is the acquisition of unobserved child nodes within an MB-unit allows the transition of SFs to CFs (Proposition 1). The second route is achieved by completing an MB-unit; acquiring an unobserved MB feature (e.g., a spouse) can move previously observed features into $C(A)$. Several examples are provided in results (section~\ref{sec:results}) that demonstrate this transition. In algorithm~\ref{alg:edfa} (line 9), $r(U_j)$ is used as a heuristic measure to find the number of potential CF features unlocked per unit cost. This is an optimistic measure of whether an unlocked $Sp$ would produce label-flipping CF features. We do not claim global optimality of the resulting acquisition sequence from the heuristic measure; similar to other greedy AFA policies, EDFA's sequence is myopic under the budget constraint.\\
A crucial factor for trustworthy algorithmic recourse is that CFs should consider directional dependencies between features, or causal relationships, so that the proposed interventions map to actions that are realizable in the real world. Even though the MB recovers the necessary and sufficient feature set it does not recover the relational dependencies between the features. Hence, we use plausibility as a proxy to generate actionable CFs, without having to recover the full causal graph, which is accepted as a suitable alternative in prior work~\cite{smyth2022few, sokolAll2025}. Two such CF generation algorithms are used in the proposed framework, namely, EDFA-default (or EDFA) and EDFA-generative. The default version produces CFs by finding similar instances in the training set and replacing feature values in those instances. Thus, it preserves plausibility by ensuring that the CFs suggest only naturally occurring changes in features. Similar approaches have already been proposed in the literature~\cite{brughmansNICEAlgorithmNearest2023, KarKugSchVal20}. The EDFA-generative method uses a Conditional Variational Autoencoder (CVAE) based approach to generate the CFs. Similar generative model based approaches are also well known to preserve plausibility~\cite{barbera2024interpretable,delserGeneratingTrustworthyCounterfactual2024, madaan2024navigating}.

\begin{algorithm}[h]
\caption{The proposed EDFA algorithm}\label{alg:edfa}
\begin{algorithmic}[1]
\Require MB-units discovered $\mathcal{A}=\{PC_j \mapsto Sp(PC_j)\}_{j=1}^{k}$, cost function $c$, budget $B$, model $f$, test instances $x$, starting feature set $a_s$ with observed values $x_{a_s}$.
\Ensure Acquired set $A$
\State $A \leftarrow a_s$,\quad $\mathit{cost} \leftarrow \sum_{X_i \in a_s} c(X_i)$
\State Compute $\alpha(\{X_i\},x,y^{*}),~\forall X_i \in MB(Y) \cap A$
\State Partition into $\mathcal{C}(A),\mathcal{S}(A),\mathcal{A}_{alt}$
\While{$\mathit{cost} < B$ \textbf{and} $\exists$ unacquired MB features}
    \If{$\exists$ partially acquired MB-unit $U_j$}
        \State $U^{*} \leftarrow U_j$ \Comment{complete $U_j$}
    \Else
        \ForAll{unstarted $U_j$ with PC member $PC_j$}
            \State $r(U_j) \leftarrow \dfrac{|Sp(PC_j)|+1}{c(PC_j)}$ \Comment{spouses unlocked per unit cost}
        \EndFor
        \State $U^{*} \leftarrow \operatorname*{arg\,max}_{j}\, r(U_j)$
    \EndIf
    \If{$PC_{j^{*}}$ of $U^{*} \notin A$}
        \State $X_{next} \leftarrow PC_{j^{*}}$ \Comment{unlock spouses by acquiring PC first}
    \Else
        \State $X_{next} \leftarrow \operatorname*{arg\,max}_{X_i \in Sp(PC_{j*}) \backslash{A}}\, \dfrac{I(Y;X_i \mid X_A = x_A)}{c(X_i)}$
    \EndIf
    \If{cost + $c(X_{next}) \leq B$}
        \State Observe $x_{next}$ for $x$
        \State $A \leftarrow A \cup \{X_{next}\}$
        \State cost=cost+$c(X_{next})$
    \Else
        \State \textbf{break}
    \EndIf
    \State Recompute $\alpha(\{X_i\},x,y^{*}),~\forall X_i \in MB(Y) \cap A$
    \State Repartition  $\mathcal{C}(A),\mathcal{S}(A),\mathcal{A}_{alt}$
\EndWhile
\end{algorithmic}
\end{algorithm}

\subsection{Recourse validity guarantees}
As mentioned in section~\ref{sec:theory}, Risk Controlling Prediction Sets with Hoeffding-Bentkus Upper Confidence Bounds (termed RCPS-HB from here onward) is used as the technique to derive validation guarantees of recourse options. RCPS-HB provides validity guarantees in the form of probabilistic bounds, which are calculated using the held-out calibration set. We then define several acquisition stopping criteria (stopping rules) for a given test instance.
\begin{enumerate}
    \item \textbf{Stop at the earliest step (baseline)} - the acquisition stops naively right after the first acquisition step.
    \item \textbf{Stop at the last step (baseline)} - Stops naively after spending the whole budget and acquiring all the features.
    \item \textbf{Stop at the first safe step (safe rule)} - Stops at the step when the uncertainty ($1-max(p)$, where $max(p)$ is the predicted top-class probability) is equal to or below the threshold calculated by RCPS-HB.     
\end{enumerate}

It was found that the baseline rules do not provide satisfactory guarantees that are robust enough. Hence, the safe rule is used throughout the reported results.

\begin{algorithm}
  \caption{RCPS--HB validation guarantee and deployment for a test instance}
  \label{alg:rcps-hb}
  \begin{algorithmic}[1]
  \Require calibration set $\mathcal{D}_{\mathrm{cal}}=\{z_i\}_{i=1}^{n}$, where each
           $z_i$ holds $m$ seed-varied EDFA trajectories for \emph{one fixed} instance;
           risk level $\alpha$; confidence $\delta$; uncertainty signal $s(\cdot)$;
           grid size $G$; test instance $x_\star$
  \Ensure  threshold $\hat\tau$ with $\Pr_{\mathcal{D}_{\mathrm{cal}}}\!\big(R(\hat\tau)\le\alpha\big)\ge 1-\delta$;
           committed step $k_\star$ for $x_\star$
  \Statex
  \Function{HBpValue}{$\hat R,\, n,\, \mu$} \Comment{Hoeffding--Bentkus tail $p$-value}
    \If{$\mu \le \hat R$} \State \Return $1$ \EndIf
    \State $k \gets \lceil n\hat R\rceil$
    \State $B \gets \sum_{j=0}^{k}\binom{n}{j}\mu^{j}(1-\mu)^{\,n-j}$ \Comment{$\mathrm{BinomCDF}(k;n,\mu)$}
    \State \Return $\min\!\big(\, e\cdot B,\;\; \exp(-2n(\mu-\hat R)^2)\,\big)$
  \EndFunction
  \Statex
  \Function{HBucb}{$\hat R,\, n,\, \delta$} \Comment{smallest mean not ruled out at level $\delta$}
    \If{$\hat R \ge 1$} \State \Return $1$ \EndIf
    \State $(\ell,\, u) \gets (\hat R,\, 1)$
    \While{$u-\ell > \mathrm{tol}$}
       \State $\mu \gets \tfrac12(\ell+u)$
       \If{\Call{HBpValue}{$\hat R, n, \mu$} $\le \delta$}
          \State $u \gets \mu$
       \Else
          \State $\ell \gets \mu$
       \EndIf
    \EndWhile
    \State \Return $u$ \Comment{$= \min\{\mu\in[\hat R,1] : p_{\mathrm{HB}}(\hat R;n,\mu)\le\delta\}$}
  \EndFunction
  \Statex
  \State \textbf{Derive the bound (calibration).}
  \State $\mathcal{P} \gets \{\, s(r) : r \in z_i,\ i\le n,\ \mathrm{cf\_cost}(r)\neq\bot \,\}$ \Comment{signal pool}
  \State $\{\tau_1<\dots<\tau_G\} \gets \textsc{Unique}\big(\textsc{Quantile}(\mathcal{P},\, \{0,\tfrac1{G-1},\dots,1\})\big)$
  \For{$j=1$ \textbf{to} $G$} \Comment{empirical risk curve $\hat R(\tau)$}
     \For{$i=1$ \textbf{to} $n$}
        \State $k_{ij} \gets$ first step $k$ of $z_i$ with $s(r_k)\le\tau_j$, else last legal step \Comment{preferred rule}
        \State $L_{ij} \gets \dfrac{1}{m}\sum_{t=1}^{m}\mathbf{1}\!\big[\neg\big(\text{cf valid at }k_{ij} \wedge \hat y_{k_{ij}}=\hat y_K\big)\big]$
               \Comment{$\bot$ if no legal step}
     \EndFor
     \State $\hat R_j \gets \mathrm{mean}\{\,L_{ij} : L_{ij}\neq\bot\,\}$
  \EndFor   
  \If{$\hat R$ is non-decreasing in $j$}
     \State $\hat R^{\mathrm{mono}} \gets \hat R$
  \Else
     \State $\hat R^{\mathrm{mono}}_j \gets \max_{j'\le j}\hat R_{j'}$ \Comment{conservative monotonization}
  \EndIf
  \For{$j=1$ \textbf{to} $G$}
     \State $U_j \gets \Call{HBucb}{\hat R^{\mathrm{mono}}_j,\, n,\, \delta}$ \Comment{$1-\delta$ UCB on $R(\tau_j)$}
  \EndFor
  \State $\mathcal{F} \gets \{\, j : U_j \le \alpha \,\}$
  \If{$\mathcal{F} = \varnothing$}
     \State $\hat\tau \gets \tau_1$,\quad \textsc{feasible}$\gets$\textbf{false} \Comment{no $\tau$ controls risk $\Rightarrow$ strictest rule}
  \Else
     \State $\hat\tau \gets \tau_{\max\mathcal{F}}$,\quad \textsc{feasible}$\gets$\textbf{true} \Comment{largest $\tau$ with UCB $\le\alpha$}
  \EndIf
  \State \textit{Guarantee:} $\Pr_{\mathcal{D}_{\mathrm{cal}}}\!\big(R(\hat\tau)\le\alpha\big)\ge 1-\delta$.
  \Statex
  \State \textbf{Deploy the certified rule on $x_\star$.}
  \State run EDFA acquisition on $x_\star \Rightarrow$ trajectory $(r_0, r_1, \dots, r_{K})$
  \State $k_\star \gets$ first step $k$ with $s(r_k)\le\hat\tau$, else last legal step \Comment{same preferred rule}
  \State \Return $\hat\tau,\ k_\star,\ \text{prediction/recourse at } r_{k_\star}$ \Comment{inherits the $1-\delta$ guarantee}
  \end{algorithmic}
  \end{algorithm}

\subsection{Experimental Setup}

We use a Feature Tokenizer Transformer (FTT) as the primary predictive model. In addition, an XGBoost (XGB) model is used in order to show how the proposed method behave with decision tree-based classifiers. The XGB model (as well as tree-based models in general) is considered a suitable model for tabular datasets, and is also regarded as even better than deep learning methods in certain scenarios~\cite{grinsztajnWhy}. FTT model is a more recently proposed deep learning architecture for tabular datasets~\cite{gorishniy2021revisiting}. A Bayesian Network (BN) model, (the optimally MB aligned model) is used as the ideal predictive model. However, we use it only on the Adult Income dataset, as it is not computationally feasible across all datasets (i.e., BN models require computing Bayesian Networks over large datasets, which is computationally demanding). To make predictions on feature subsets (during the feature acquisition process), we implement and train predictive models on those subsets. For the XGB model, a separate classifier is trained for each subset of features on training dataset. The FTT model was trained as only single models with binary acquisition masks, which is considered as a standard technique. \\
A standard feature preprocessing pipeline is used where continuous features are standard normalized and categorical features are label encoded. For the time-varying features, summary statistics (mean, min, max, last value) are used. Certain features that have high ratio of missing values, with very high cardinality or have constant values for all entries are dropped as well (e.g., diabetes130 dataset). MB discovery and training is carried out using the train set. At each acquisition step, the trained model is used to generate explanations for randomly sampled instances from the test set. We repeat this sampling 10 times to obtain explanations and their corresponding evaluations, averaged over the 10 runs.\\
For the convenience of running the experiments, two phases are used. In phase 1 of the experiments, the behavior of the recourse frontier during the feature acquisition process is primarily evaluated. The EDFA algorithm~\ref{alg:edfa} is tested with three different MB discovery methods; EAMB, HITON-MB and BAMB~\cite{lingBAMB2019}. 
3 SOTA AFA methods from the literature (DIME~\cite{gadgil2024estimating}, EDDI~\cite{maEDDI2019} and GSMRL~\cite{liActive2021}) and one random MB feature selection method are used as the baselines. Three well-established counterfactual recourse search methods from the literature are selected as baseline methods to evaluate the validation guarantees of the EDFA algorithm. Validation guarantees are evaluated in phase 2 of the experiments. Different CF search methods picked from the literature are utilized (namely; DiCE~\cite{dice}, NICE~\cite{brughmansNICEAlgorithmNearest2023}, and PROBE~\cite{probe}) within the EDFA algorithm, including the default and generative methods proposed originally in the EDFA framework. The selection of the aforementioned CF search methods are based on whether they provide plausible CFs, which is crucial to ensure that the CFs are realistic.  

\subsubsection{Datasets}
We pick 7 tabular datasets from two domains (finance and healthcare) that provides a diverse range of dataset sizes and features. A summary of the datasets are provided in table~\ref{tab:datasets} with the type of task that each dataset is aimed at. The Adult Income dataset~\cite{adult_uci} contains records census data of of a person and labels if the person's annual income exceeds \$50K or not. The German Credit dataset contains loan applications and labels good vs. bad credit risk of the applicants~\cite{hofmann1994german}. Similarly, HELOC dataset labels if a home equity credit applicant would repay within two years or not (credit risk)~\cite{fico2018heloc}, Taiwan Credit dataset labels whether credit card holder would default next month based on their billing and payment history~\cite{yeh2009taiwan}, and the GMSC dataset labels if a borrower experiences financial distress within two years~\cite{gmsc2011kaggle}. The ACS Income dataset is similar to the Adult dataset (census based income)~\cite{ding2021folktables}. Diabetes130, one of the two clinical setting datasets, labels if a diabetic patient would be readmitted or not~\cite{strack2014diabetes130}. \\
In order to manage the large number of experiments  spanning multiple configurations (i.e., baseline methods, MB discovery etc.) across the datasets, the dataset size is capped at 50K. The datasets that were subsampled (Diabetes130) were sampled using a stratified approach; preserving the class distributions. The implications of this choice are further discussed in the discussion (section~\ref{sec:discussion}). Table~\ref{tab:datasets2} presents the starting feature set, the cost assignments, and the sensitive features deemed for each dataset. The cost assignment was done by the authors based on reasonable assumptions on what the perceived cost would be in practical scenarios (table~\ref{tab:datasets2}. (similar approaches have been followed in the literature, e.g., assigning fictitious costs~\cite{hoarau2026}). All the datasets were obtained from OpenML data repository~\footnote{https://www.openml.org/}. 


\begin{table}[h]
\centering
\caption{Dataset statistics showing train/test splits, number of features, and prediction targets.}
\label{tab:datasets}
\begin{tabular}{|c|c|c|}
    \hline
     Dataset & \# of features & \# Target\\
     \hline
     German Credit & 20 & Credit risk\\
     Adult Income & 13 & Income $>$50K\\
     HELOC & 22 & Credit risk\\
     Taiwan Credit  & 23 & Credit default\\
     ACS Income & 10 & Income $>$ 50K\\
     GMSC & 10 & Financial distress \\
     Diabetes130 & 39 & Hospital Readmission\\
     \hline
\end{tabular}
\end{table}

\subsubsection{Evaluation Metrics}
We pick evaluation metrics that measure the changes of available recourse at each acquisition step. Accuracy is used for evaluate the predictive performance of the models under the acquisitions. The following evaluation metrics are predominantly used in the results reported.  `Normalized Cost' was selected to gauge the position of the feature acquisition process instead of the `number of features acquired' usually used in the AFA literature. This choice emphasizes the cost-recourse tradeoff across methods, which is more applicable to real-world recourse settings where cost reflects the effort/resources spent acquiring features rather than their count. In other words, this represents the average `cost of acquiring the features' over the test set. The metrics shown below produce `trajectories' depicting how each method behaves in terms of the number of available recourse options at each acquisition step. 

\begin{enumerate}
    \item Recourse growth vs. Normalized cost - This measures the number of valid counterfactual recourse options at a given feature acquisition cost. The expected ideal behavior of this metric is to increase as the cost accumulates.
    \item Strict Fidelity SF count vs. Normalized cost - Provides the count of semifactual features at a given feature acquisition cost. This metric should decrease as the cost accumulates.
    \item Predictive accuracy - The predictive accuracy over the sampled test set, at a given acquisition cost.
    \item $|C(A)|/|MB|$ - The ratio of counterfactual active features to the size of the MB. This quantifies the number of acquired CF features (or, features that would change the prediction if perturbed) as a ratio to the total number of features in the MB. This value should also increase with the accumulated cost as the size of $|C(A)|$ only grows when CF features are acquired.
\end{enumerate}


The `CF Valid@K' metric is used to measure the validation guarantees. The primary intuition behind this metric is that; "given a valid recourse proposed early (with partial information, to save acquisition cost), is it still valid once all the information is acquired?". It is calculated in the steps provided below.

\begin{enumerate}
    \item Calculate the losses across the instances in the calibration set, and obtain the mean value ($\hat R(\tau)$, here $\tau$ corresponds to a threshold derived based on the confidence of the prediction of any instance, at a given AFA stage)~\ref{eq:prob1}.
    \item Then the metric is evaluated as follows; CF Valid@K = $1-\hat R(\tau)$.
\end{enumerate}
In order to evaluate the quality of the generated CFs, three widely used CF evaluation metrics in the literature are used; L0 distance (measures sparsity-the number of feature changes), L2 distance (proximity to the original datapoint), and plausibility (whether the CF is an inlier of the training data distribution). To measure the plausibility in Table~\ref{tab:cfquality}, a Local Outlier Factor (LOF) model was used~\cite{LOF}. LOF calculates the plausibility of the CFs by quantifying how close they are to the data instances in the original dataset. All four methods use the same (default) CF search procedure, so the differences reflect which features each method acquired rather than which CF searcher was used.
$\delta=0.05 ($eq~\ref{eq:bound0}) is used as the error rate throughout the experiments. Two invalidation rates ($\alpha\in\{0.2,0.3\}$) are used, which provide two risk levels (where $\alpha=0.3$ depicts a looser risk level) to carry out the experiments.

\section{Results}
\label{sec:results}
\subsection{Phase 1 - Counterfactual recourse growth}
\begin{figure}[htbp]
    \begin{subfigure}{\linewidth}
        \centering
        \includegraphics[width=\linewidth]{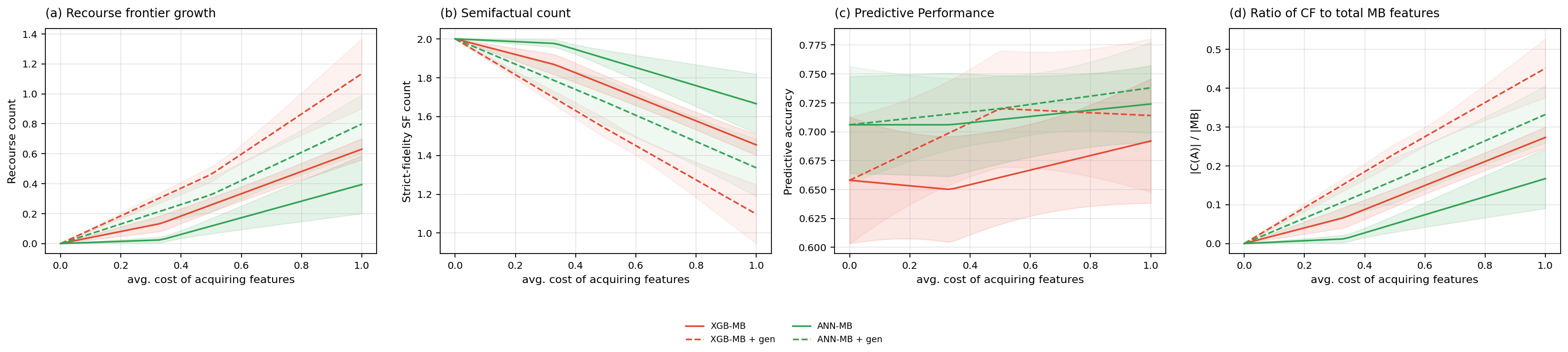}
        \caption{German Credit}
        \label{fig:german1}
    \end{subfigure}

    \begin{subfigure}{\linewidth}
        \centering
        \includegraphics[width=\linewidth]{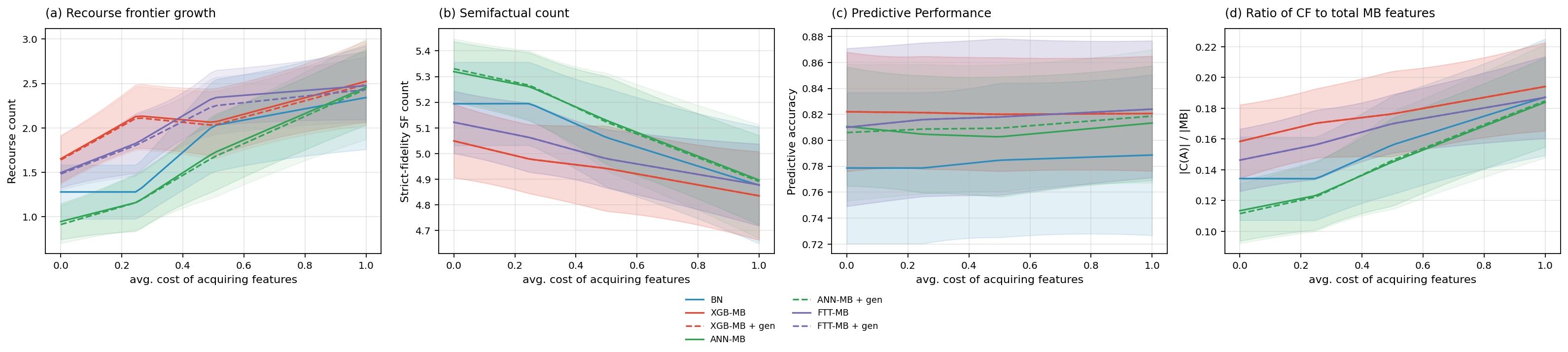}
        \caption{Adult Income}
        \label{fig:adult1}
    \end{subfigure}

    \begin{subfigure}{\linewidth}
        \centering
        \includegraphics[width=\linewidth]{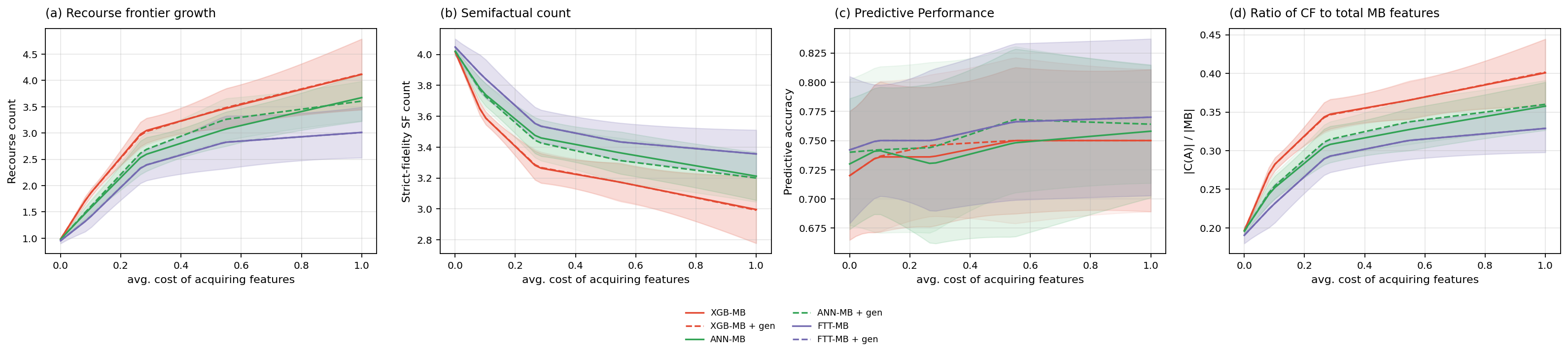}
        \caption{HELOC}
        \label{fig:heloc1}
    \end{subfigure}

    \caption{The figures shows the recourse frontier size and the accuracy at each step of the feature acquisition for the EDFA method. All the subplots from figures (a) - (d) have the avg. cost of acquiring features which is measured by normalized cumulative costs calculated from the sampled test set. The dotted lines correspond to the HITON and BAMB MB discovery and the solid lines to the EAMB method. The bands represent the standard deviation measurements obtained from multiple runs. The increasing trend in sub figures (a) and (d) correspond to the growth of the recourse frontier as more features are acquired. In tandem, the SFs are decreased as shown in the subfigures in (b). The accuracy remains stable throughout the acquisition process as shown in sub figure (c).}
    \label{fig:res1}
\end{figure}

\begin{figure}[htbp]
 \begin{subfigure}{\linewidth}
        \centering
        \includegraphics[width=\linewidth]{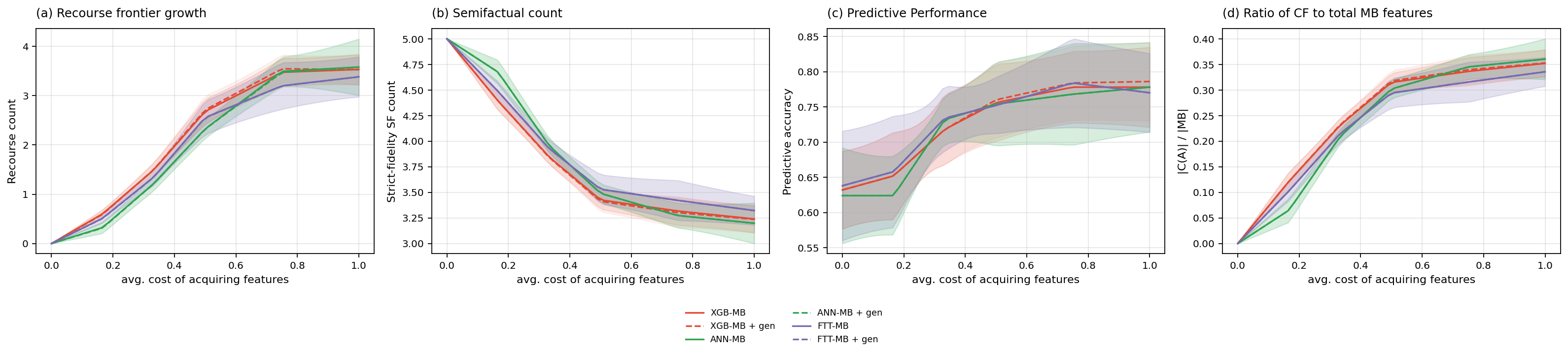}
        \caption{ACS Income}
        \label{fig:acs1}
    \end{subfigure}

    \begin{subfigure}{\linewidth}
        \centering
        \includegraphics[width=\linewidth]{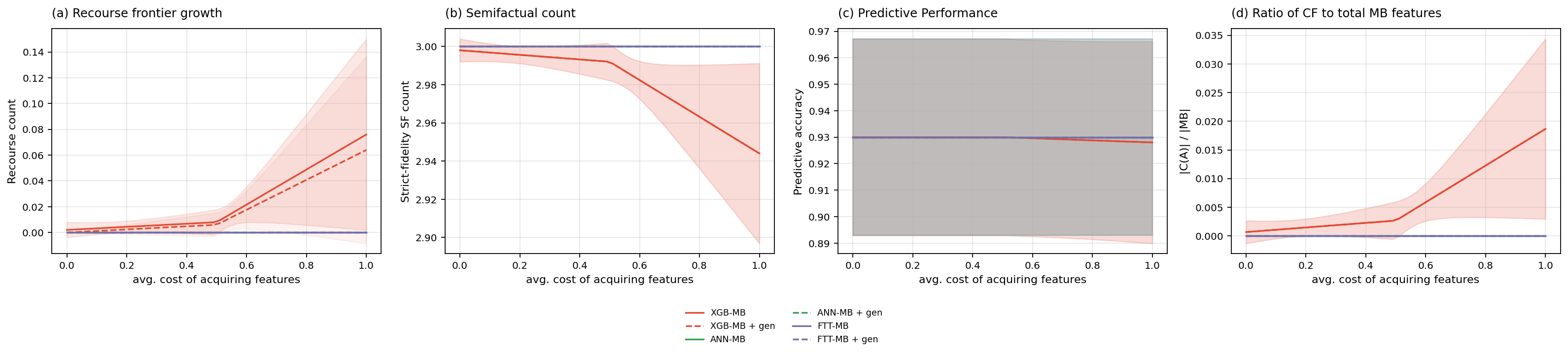}
        \caption{GMSC}
        \label{fig:gmsc1}
    \end{subfigure}

    \begin{subfigure}{\linewidth}
        \centering
        \includegraphics[width=\linewidth]{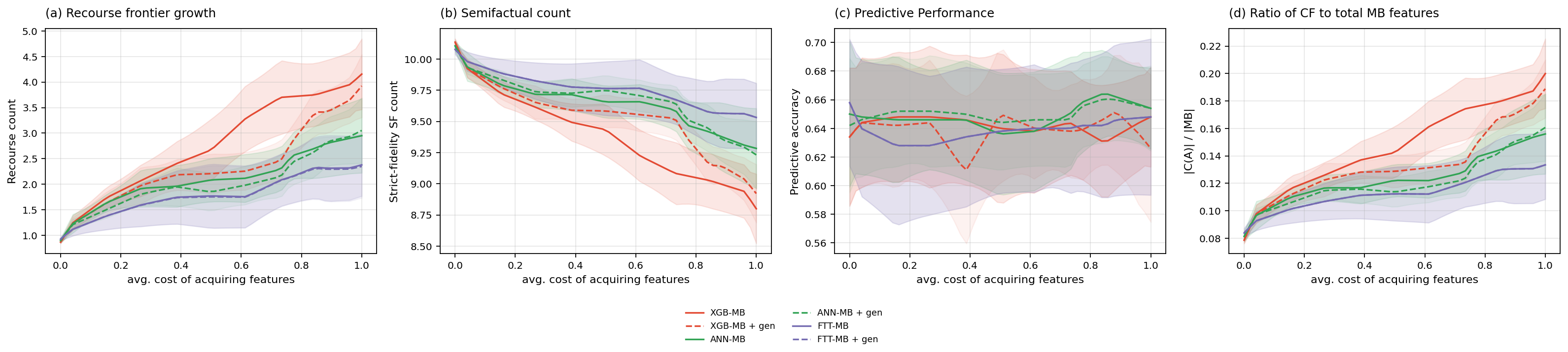}
        \caption{Diabetes}
        \label{fig:diabetes1}
    \end{subfigure}

    \caption{Recourse frontier change and accuracy at each step of the feature acquisition for the EDFA method for ACS Income, GMSC and Diabetes datasets. The plots style is the same as in figure~\ref{fig:res1}. The same pattern from figure~\ref{fig:res1} can be seen here as well, where EDFA method consistently acquires features that can provide recourse while maintaining an acceptable level of predictive performance. The exception is the GMSC dataset.}
    \label{fig:res2}
\end{figure}

Tables~\ref{tab:afacomp1} and~\ref{tab:cfquality} show the average number of features acquired by each AFA method throughout the trajectories, and the qualities of the CFs generated throughout the trajectories. The proposed EDFA method picks the least number of features on average to generate CFs (Table~\ref{tab:afacomp1}). The generated CFs are returned as the most sparse (lower L0 distance), and plausible (percentage of inlier CFs) while being closer to the original instances (lower L2 distance) for most scenarios. For instance, diabetes130 shows the best performance of EDFA in terms of CF quality, compared to other AFA methods.\\ Figures~\ref{fig:res1}-\ref{fig:res2} present how the EDFA acquisition method influences the recourse frontiers of the datasets with its two MB discovery variants. The sub figures show the growth of the recourse frontier size, the decline of semifactuals, the predictive performance of the model and the ratio of the CF figures to the size of the MB ($|C(A)|/|MB|$), respectively. The general pattern is that the recourse frontier grows as more features are acquired (at higher cost) while maintaining predictive performance (accuracy) at an acceptable level.

\begin{table}[ht]
    \centering
    \begin{tabular}{|c|c|c|c|c|c|c|}
    \hline
     Dataset&Features&MB size&EDFA&DIME&EDDI&GSMRL\\
     \hline
     Adult Income&13&6&\textbf{3}&8&8&7.6\\
     German Credit&20&2&\textbf{2}&15.9&15.7&15.1\\
     HELOC&22&5&\textbf{4}&18&17.9&17.6\\
     Taiwan Credit&23&7&\textbf{7}&18&16.2&18\\
     ACS Income&10&5&\textbf{5}&7&7&7\\
     GMSC&10&3&\textbf{2}&7&6.3&7\\
     diabetes130&39&11&\textbf{10}&29&29.6&29.2\\
     \hline
\end{tabular}
    \caption{The average number of features acquired by each AFA method on the datasets. EDFA uses the default CF search method here. EDFA picks the least number of features compared to the baseline methods, as it is inherently limited to the MB. Yet, it maintains comparable accuracy with baseline methods with lesser number of features acquired (Figures~\ref{fig:res1}-\ref{fig:res2}).}
    \label{tab:afacomp1}
\end{table}

\begin{table}[htbp]
    \centering
    \small
    \begin{tabular}{|c|c|c|c|c|}
    \hline
     Dataset&AFA method&avg. $L_0$&avg. $L_2$&Plausibility (\%)\\
     \hline
     \multirow{4}{*}{Adult Income}&EDFA&\textbf{1.58}&\textbf{0.191}&79.4\\
     &DIME&2.43&0.238&79.9\\
     &EDDI&2.42&0.245&\textbf{80.2}\\
     &GSMRL&2.41&0.244&79.8\\
     \hline
     \multirow{4}{*}{German Credit}&EDFA&\textbf{1.10}&0.379&\textbf{84.7}\\
     &DIME&2.09&0.254&81.5\\
     &EDDI&2.08&\textbf{0.242}&81.8\\
     &GSMRL&2.12&0.248&80.5\\
     \hline
     \multirow{4}{*}{HELOC}&EDFA&\textbf{1.57}&\textbf{0.112}&\textbf{63.1}\\
     &DIME&2.08&0.164&62.1\\
     &EDDI&2.09&0.166&59.9\\
     &GSMRL&2.09&0.169&61.4\\
     \hline
     \multirow{4}{*}{Taiwan Credit}&EDFA&\textbf{1.25}&\textbf{0.168}&\textbf{87.3}\\
     &DIME&2.18&0.142&84.8\\
     &EDDI&2.19&0.165&83.7\\
     &GSMRL&2.16&0.145&86.7\\
     \hline
     \multirow{4}{*}{ACS Income}&EDFA&\textbf{1.52}&0.178&88.5\\
     &DIME&2.08&\textbf{0.151}&87.2\\
     &EDDI&2.10&\textbf{0.151}&\textbf{88.6}\\
     &GSMRL&2.06&0.150&87.7\\
     \hline
     \multirow{4}{*}{GMSC}&EDFA&---&---&---\\
     &DIME&\textbf{2.58}&\textbf{0.130}&45.8\\
     &EDDI&2.61&0.163&33.1\\
     &GSMRL&2.61&0.118&41.8\\
     \hline
     \multirow{4}{*}{diabetes130}&EDFA&\textbf{1.45}&\textbf{0.162}&\textbf{85.0}\\
     &DIME&1.95&0.179&80.0\\
     &EDDI&2.01&0.186&79.1\\
     &GSMRL&2.00&0.186&78.6\\
     \hline
\end{tabular}
    \caption{Quality of the counterfactual recourse generated along the acquisition trajectories of each AFA method, under the same configuration as table~\ref{tab:afacomp1}. $L_0$ is the number of features a CF asks the user to change, $L_2$ is the range-normalized Euclidean distance from the factual instance, and Plausibility is the percentage of CFs that a LOF model labels as inliers. Each value averages over every counterfactual in the recourse frontier at every acquisition step, over 100 test instances per dataset (251{,}612 counterfactuals in total). EDFA yields only too few CFs to average, hence the missing entries. EDFA method generates sparse, and plausible CFs while being closer to the original data points (on average).}
    \label{tab:cfquality}
\end{table}

The next set of figures shows the comparison with other feature acquisition methods, for the XGB and FTT predictive models. We include only the results obtained with the XGB and FTT models in the plots; for brevity, we omit the remaining results. The main reason is that FTT is an advanced form of an FNN, which is sufficient to represent FNNs in general. All 3 EDFA variants (default, generative and random) are included, alongside the AFA baselines. The overarching behavior is that the recourse frontier grows as more features are acquired (more budget is spent/more cost). A stark observation is that the three SOTA baseline AFA methods provide more recourse options, as can be seen with the rapid growth of the recourse count plots compared to EDFA (subfigure (a) of each figure). However, this improvement is attributed to the reliance on non-MB features, where SOTA AFA methods keep picking features outside the MB (subfigure (d) of each figure). The FTT results for the German Credit dataset were not obtained due to the inadequacy of the dataset size, and for the same reason, they were not included in this cohort of plots.\\
Figures~\ref{fig:transitions} provide four recourse transitions extracted from the experiments. They demonstrate how explanation driven acquisition of features provide recourse options via CF explanations. In particular, they demonstrate how instances which only had SFs with limited information would be able to obtain CFs through feature acquisition. These are instances of latent recourse (Definition 9). These examples correspond to instances where MB-unit completion through $Sp$ feature acquisition provides with CFs. The first sub-figure shows one such transition observed in the Adult Income dataset. It has the starting feature set which is predicted label 0 (<50K income) by the model. With the current feature set, the instance cannot obtain CFs (only SFs). The EDFA algorithm acquires a new feature 'workclass' which is also a $Sp$ feature in the MB discovered. This enables to perturb the previous `education-num' feature and flip the model prediction owing to the MB unit completion. In this instance the perturbation of changing its value from `HS-grad' to 'Bachelor's' produce a CF, which is an actionable recourse. The cost of the acquired feature is in the same cost tier of the previous features for this instance (Table~\ref{tab:datasets2}).
The second scenario shows an instance from the Taiwan Credit dataset, where `pay\_amt1'(the most recent payment amount) is acquired. This acquisition enables to determine that the previously acquired feature `pay\_2' (indicates delayed payment) can be changed to `duly\_paid' in order to flip the original prediction. From a practical point of view, this means that the model makes the decision based on the availability of information on the most recent payments. The historical payment information alone cannot move the prediction. The cost of the acquired feature is a comparatively cheaper than some of the previously acquired features which were present due to the $PC$-first acquisition policy of EDFA.
This scenario depicts an instance where the acquisition of `num\_lab\_procedures' (number of lab tests) enables a decision on `num\_procedures' (non-lab procedures) by changing it from 0 to 1, which also aligns with practical diagnosis settings. The original prediction for this instance was `readmit' (label 1) which can now be flipped to 0, with the availability of the recourse. The final scenario shows an instance from the HELOC dataset, where `NumSatisfactoryTrades' (number of trades that are in good standing) becomes active CF feature, once the costly, `NumTrades60Ever2DerogPubRec' (number of trades/deregatory public records that have been 60+ days past due) $Sp$ feature is acquired. With the knowledge of the past derogatory trade amount, a recourse is now provided where the user is suggested to increase good trades from 39 to 42, in order to flip the prediction from label 0 (`Bad/Risky') to label 1. Until this new acquisition, `NumSatisfactoryTrades' only remains as a SF feature without the ability to flip the prediction.
\begin{figure}[htbp]

\begin{subfigure}{\linewidth}
        \centering
        \includegraphics[width=\linewidth]{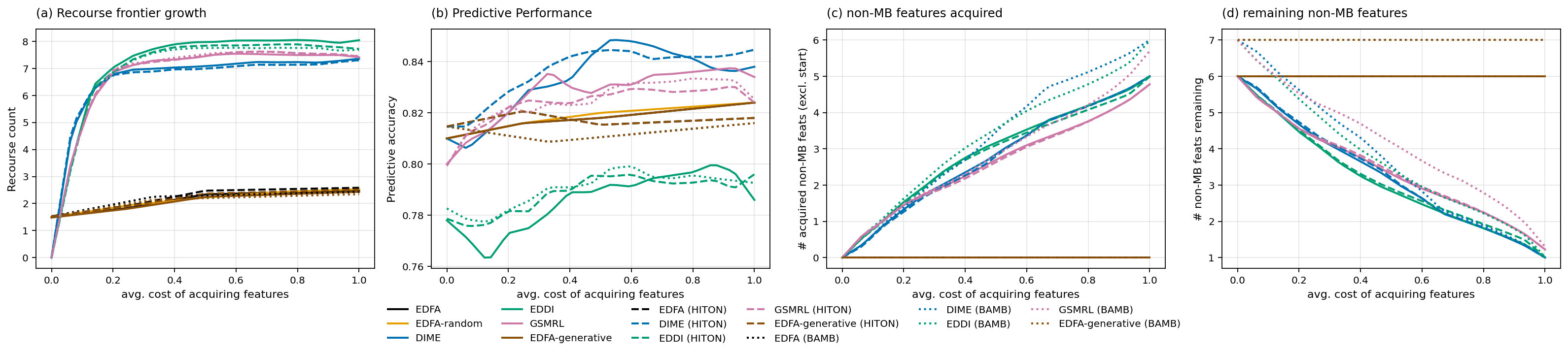}
        \caption{Adult Income - FTT}
        \label{fig:adult22}
    \end{subfigure} 
    
 \begin{subfigure}{\linewidth}
        \centering
        \includegraphics[width=\linewidth]{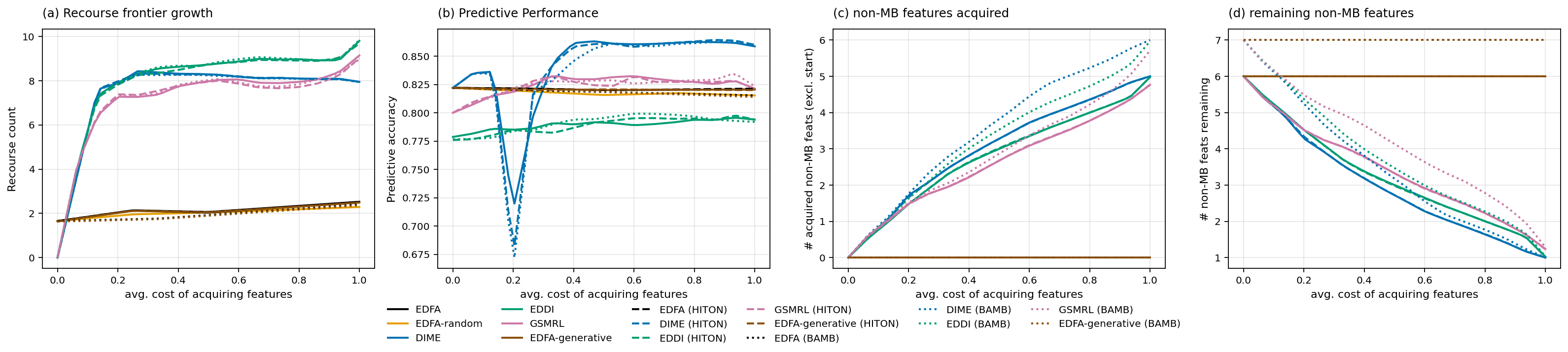}
        \caption{Adult Income - XGB}
        \label{fig:adult21}
    \end{subfigure}

    \caption{The figures show the results comparing the  EDFA method with other AFA techniques on the Adult Income dataset.(i) - results obtained with FTT as the predictive model, (ii) - results obtained with XGB as the predictive model. The 3 baseline AFA methods (DIME, EDDI, GSMRL) have dotted line plots for each, showing how they behave with reference to different MBs discovered. The baseline AFA methods provides more recourse options (CFs) than the EDFA method (a). However, they increasingly rely on non-MB features to do so as shown in the subfigures (c) and (d). The accuracy is maintained at satisfactory level by EDFA, DIME and GSMRL. EDDI lags behind in terms of the predictive performance.}
\end{figure}
\begin{figure}[htbp]
    \begin{subfigure}{\linewidth}
        \centering
        \includegraphics[width=\linewidth]{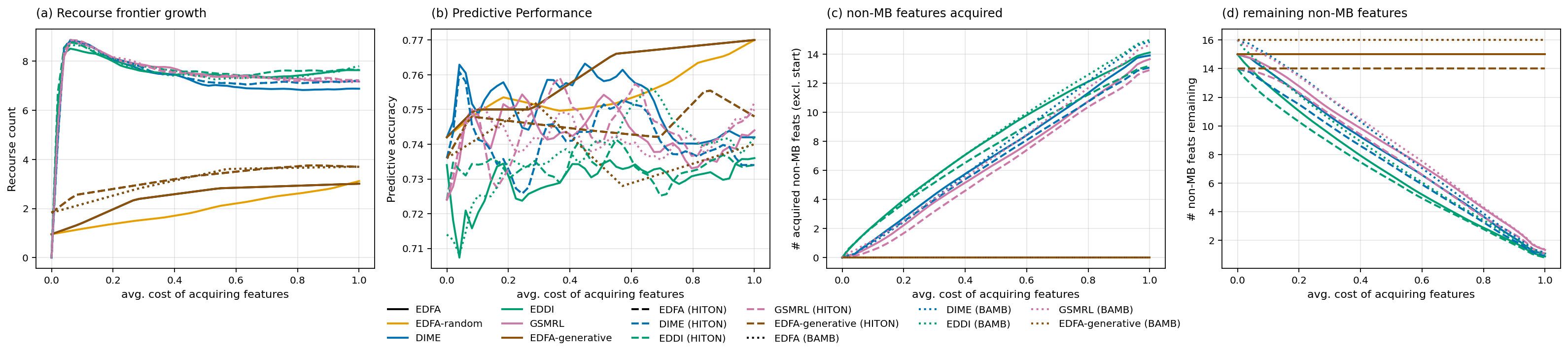}
        \caption{HELOC - FTT}
        \label{fig:heloc22}
    \end{subfigure}

     \begin{subfigure}{\linewidth}
        \centering
        \includegraphics[width=\linewidth]{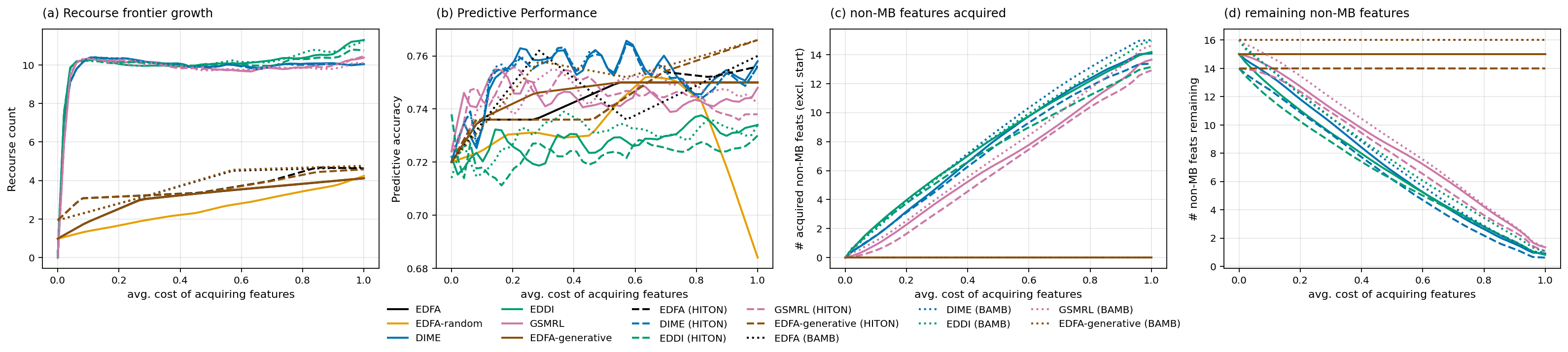}
        \caption{HELOC - XGB}
        \label{fig:heloc21}
    \end{subfigure}
    \caption{The comparisons between AFA methods on the HELOC dataset. The pattern observed in figure~\ref{fig:adult1} is present here as well, where the baseline AFA methods produce more recourse, but relies on non-MB features (shown in the sub figures (c) and (d)) in both rows).}
\end{figure}

\begin{figure}[htbp]
    \begin{subfigure}{\linewidth}
        \centering
        \includegraphics[width=\linewidth]{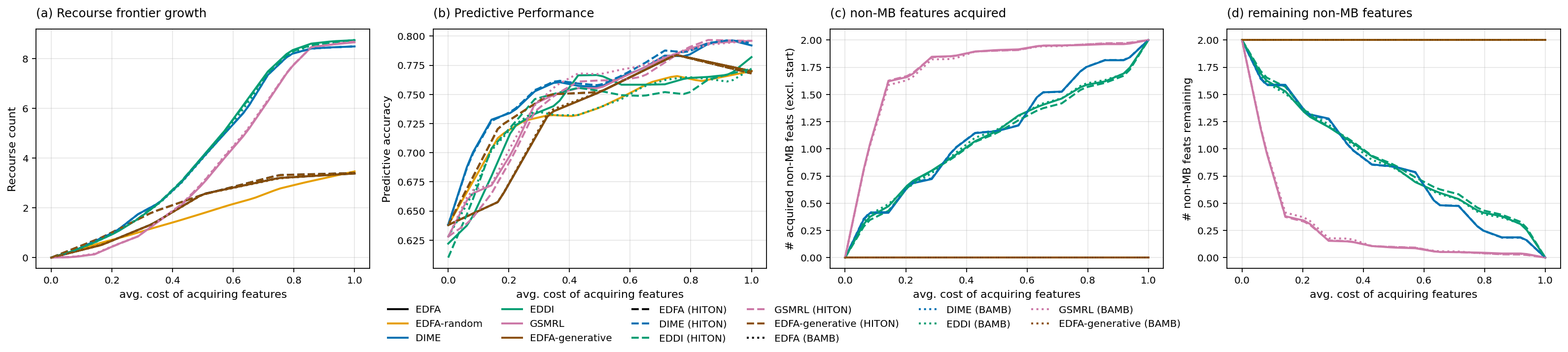}
        \caption{ACS Income - FTT}
        \label{fig:acs22}
    \end{subfigure}

     \begin{subfigure}{\linewidth}
        \centering
        \includegraphics[width=\linewidth]{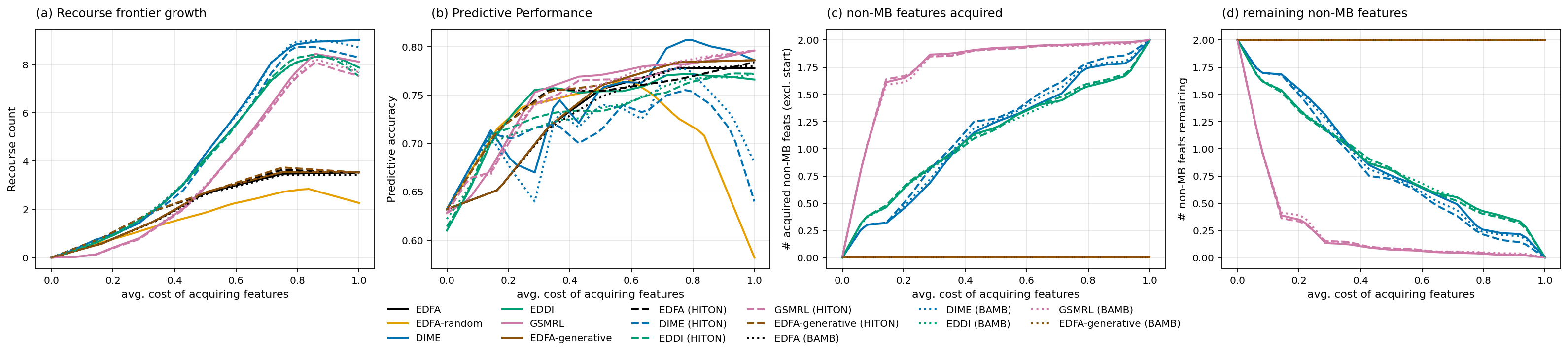}
        \caption{ACS Income - XGB}
        \label{fig:acs21}
    \end{subfigure}

    \caption{The comparisons between AFA methods on the ACS Income dataset. The baseline methods are similar in terms of predictive performance. The pattern on providing more recourse options is also present.}
\end{figure}

\begin{figure}[htbp]
   \begin{subfigure}{\linewidth}
        \centering
        \includegraphics[width=\linewidth]{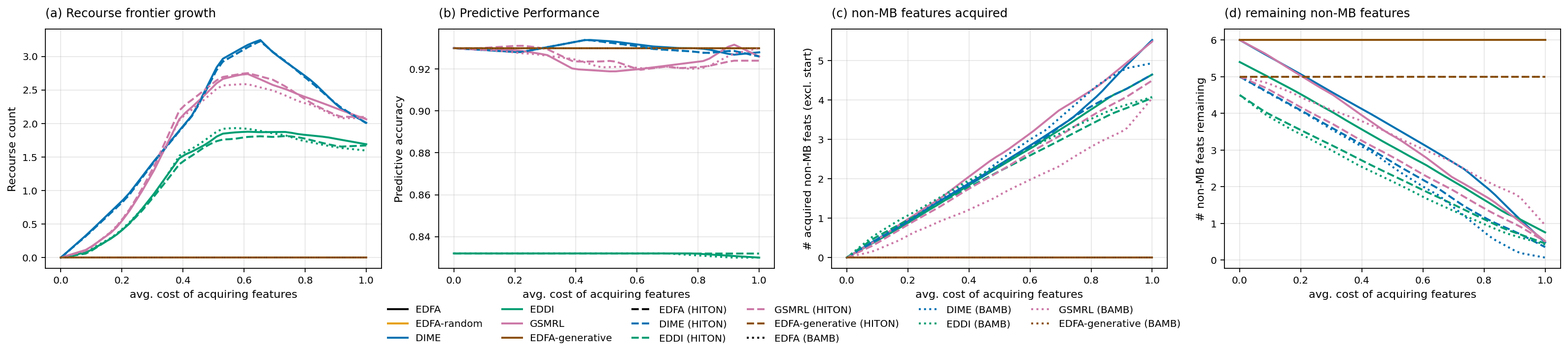}
        \caption{GMSC - FTT}
        \label{fig:gmsc22}
    \end{subfigure}

    \begin{subfigure}{\linewidth}
        \centering
        \includegraphics[width=\linewidth]{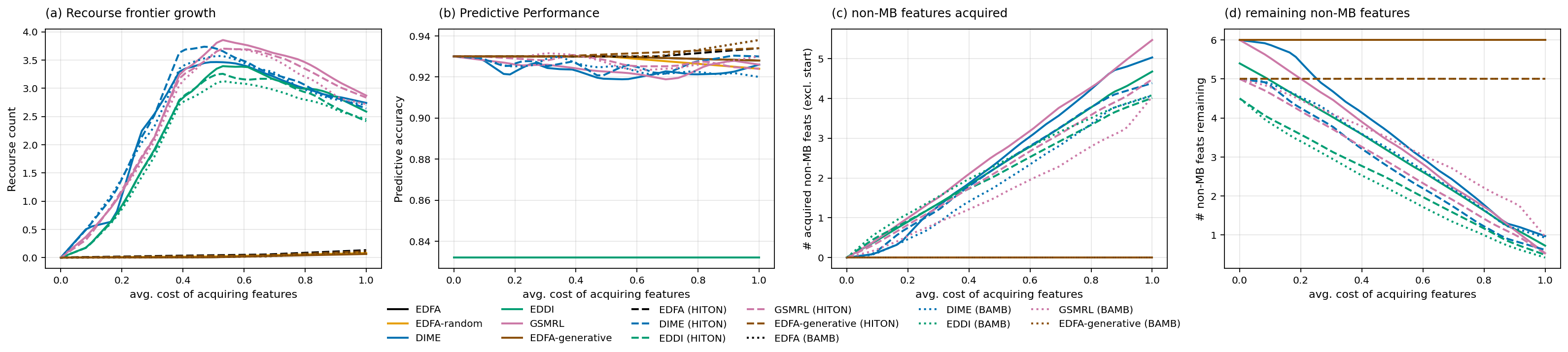}
        \caption{GMSC - XGB}
        \label{fig:gmsc21}
    \end{subfigure}
    \caption{The comparisons between AFA methods on the GMSC dataset. EDFA method fails to generate recourse options for this dataset.}
\end{figure}

\begin{figure}[htbp]
    \begin{subfigure}{\linewidth}
        \centering
        \includegraphics[width=\linewidth]{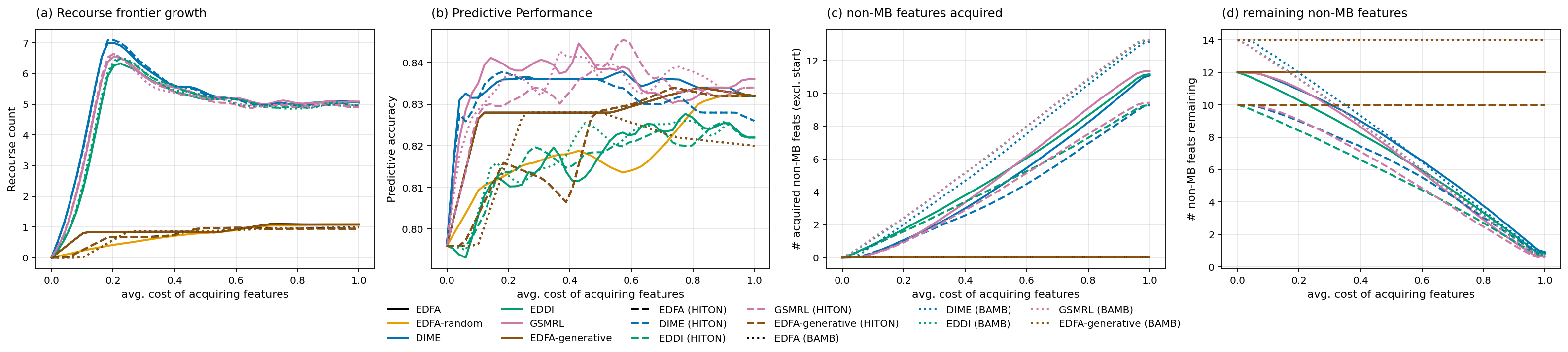}
        \caption{Taiwan Credit- FTT}
        \label{fig:taiwan22}
    \end{subfigure}

     \begin{subfigure}{\linewidth}
        \centering
        \includegraphics[width=\linewidth]{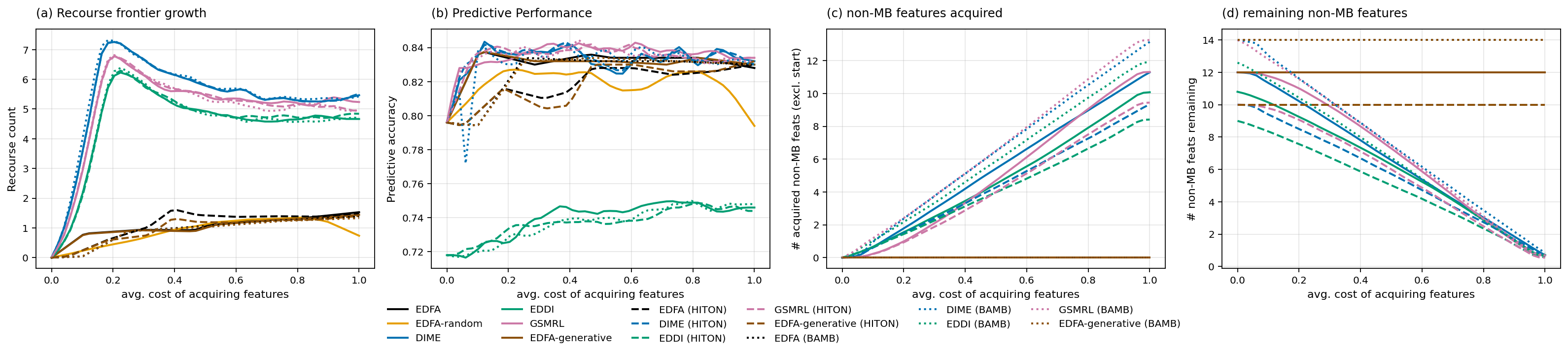}
        \caption{Taiwan Credit - XGB}
        \label{fig:taiwan21}
    \end{subfigure}

    \caption{The comparisons between AFA methods on the Taiwan Credit dataset. The EDFA shows similar predictive performance (refer subfigures in column (b))} to DIME and GSMRL, while EDDI notably lags behind.
\end{figure}

\begin{figure}[htbp]
   \begin{subfigure}{\linewidth}
        \centering
        \includegraphics[width=\linewidth]{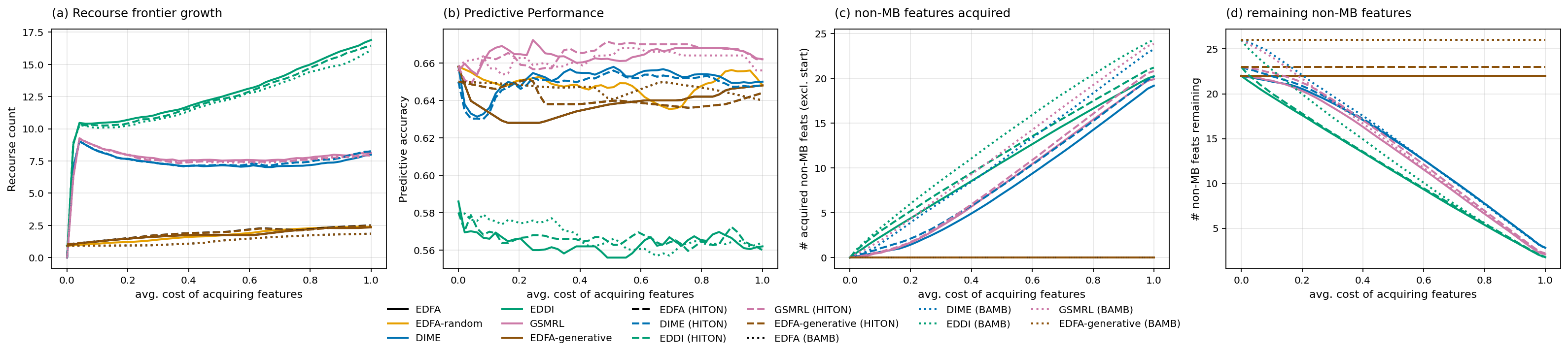}
        \caption{Diabetes - FTT}
        \label{fig:diabetes22}
    \end{subfigure}

     \begin{subfigure}{\linewidth}
        \centering
        \includegraphics[width=\linewidth]{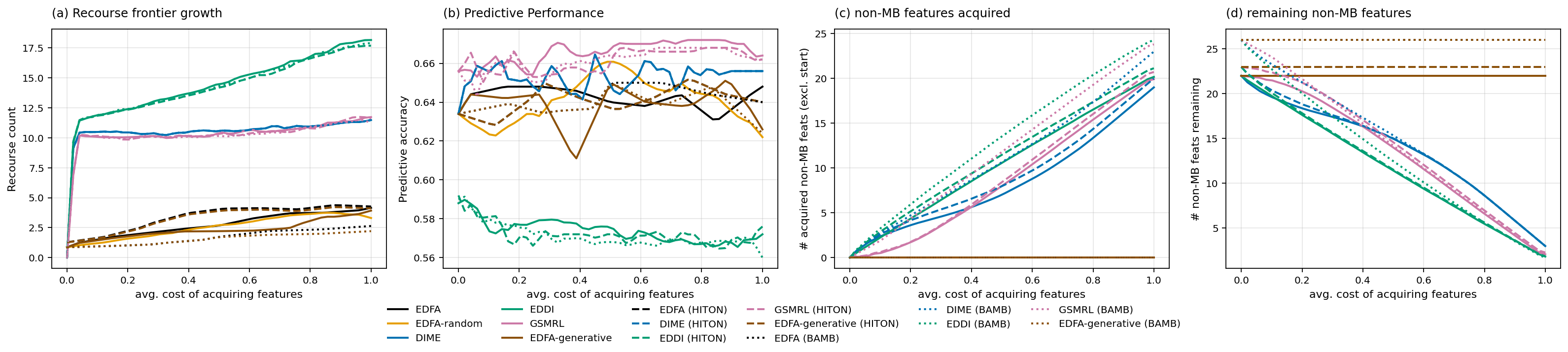}
        \caption{Diabetes - XGB}
        \label{fig:diabetes21}
    \end{subfigure}
    \caption{The comparisons between AFA methods on the Diabetes dataset. EDDI AFA method provides more recourse options with the help of non-MB features but has lower predictive performance.}
\end{figure}



\subsection{Phase 2 - Validation Guarantees}
\begin{figure}
    \centering
    \includegraphics[width=\linewidth]{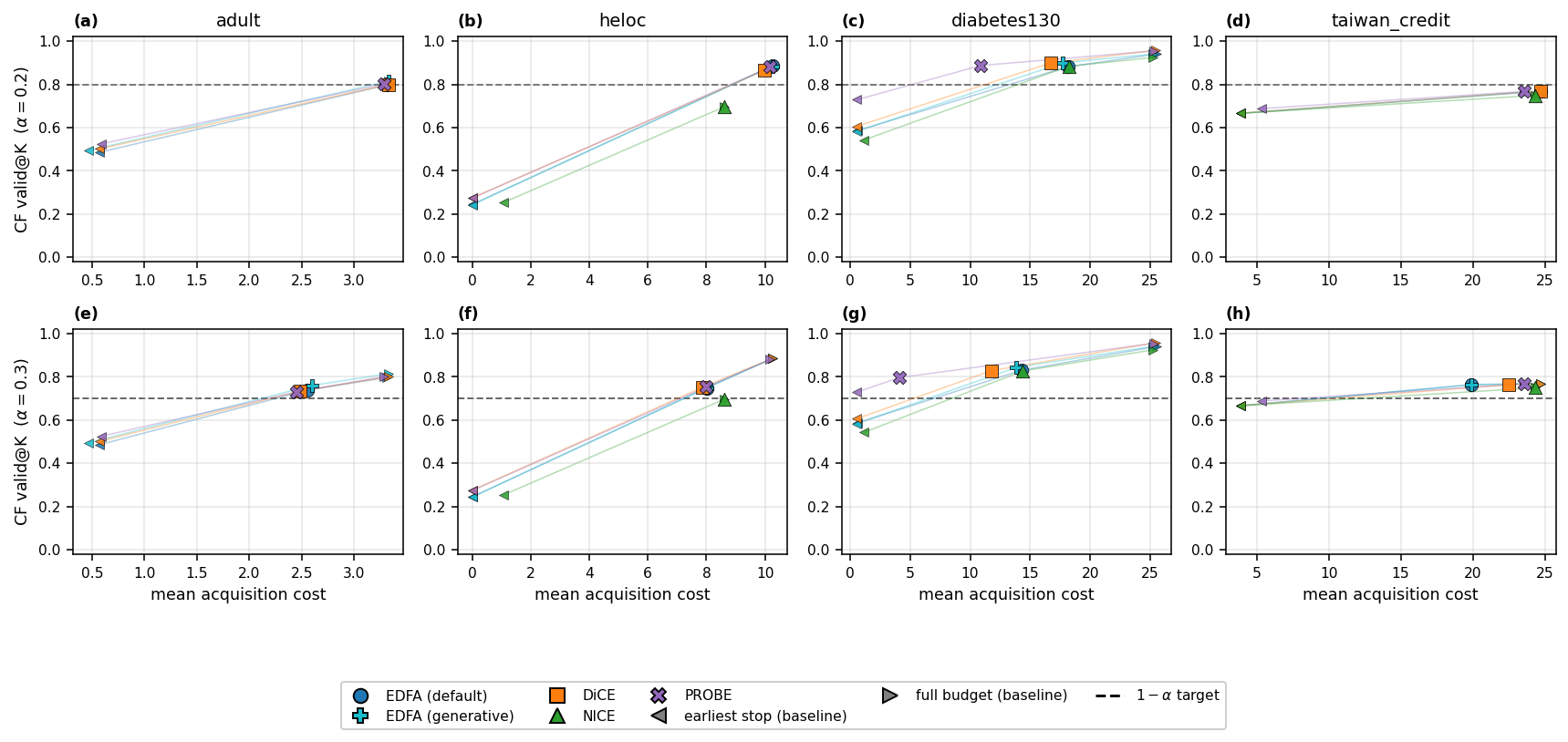}
    \caption{The probabilistic guarantee on validation rate (CF Validation@K) for $\alpha=0.2$ (top row) and $\alpha=0.3$ (bottom row) for the XGB model. These $\alpha$ levels are represented in dotted horizontal lines. The earliest stop rule and exhausting full budget rule are included as baselines~\ref{sec:method}. The instances that cross the $alpha$ line earlier provide lower cost recourse options that are also guaranteed to be valid at even with the full feature set. The safe stop rule (corresponds to the mid points on each trace) provides expected guarantee without spending the full budget. EDFA, PROBE and DiCE CF search methods are able to support this validity guarantee in most scenarios. NICE fails even with full feature budget, for the HELOC dataset, whereas PROBE offers lower cost recourse options for the Diabetes130 dataset.}
    \label{fig:g1}
\end{figure}

\begin{figure}
    \centering
    \includegraphics[width=\linewidth]{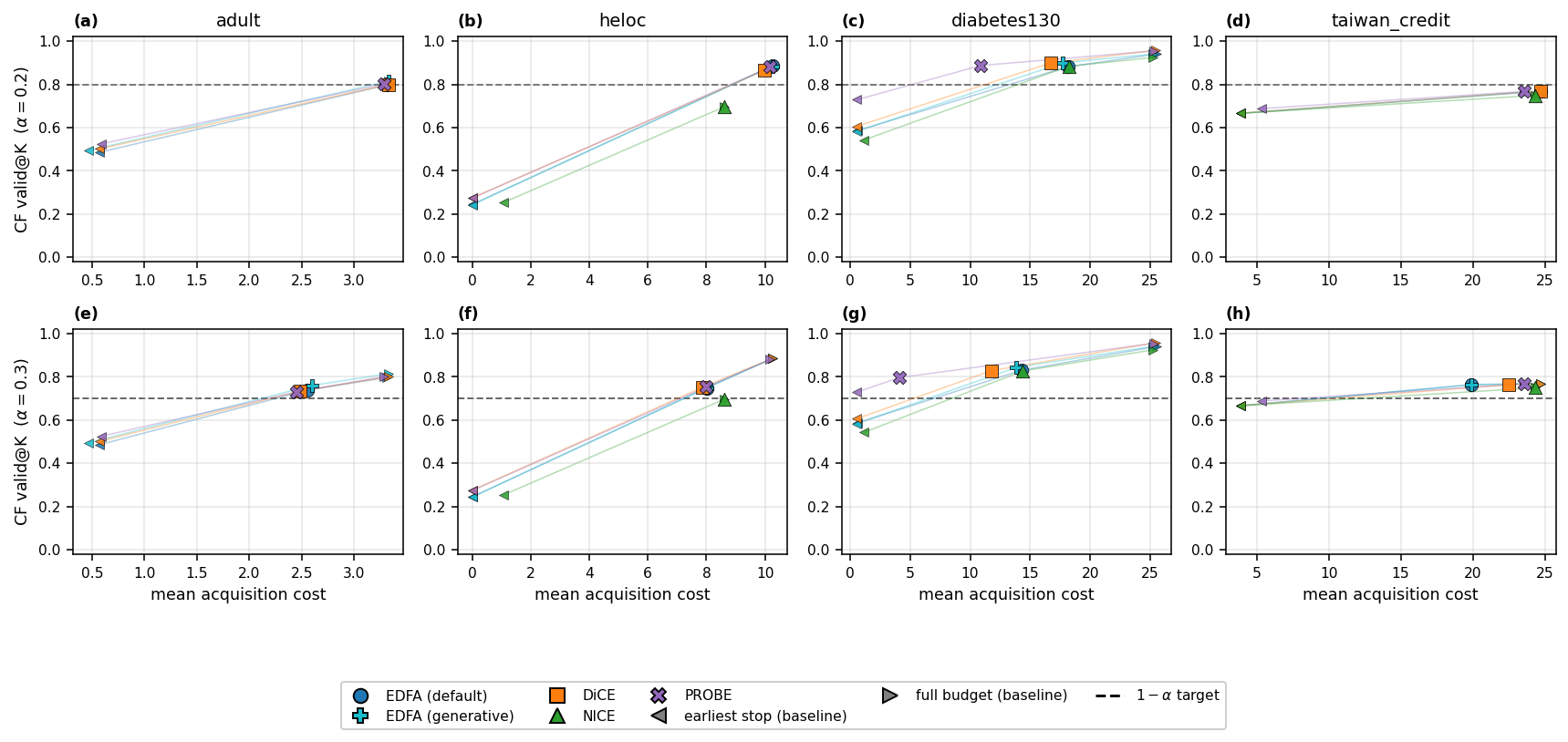}
    \caption{The probabilistic guarantee on validation rate (CF Validation@K) for $\alpha=0.2$ (top row) and $\alpha=0.3$ (bottom row) for the FTT model. The results are almost similar to the XGB model (figure~\ref{fig:g1}).}
    \label{fig:g2}
\end{figure}

\begin{figure}
    \centering
    \includegraphics[width=\linewidth]{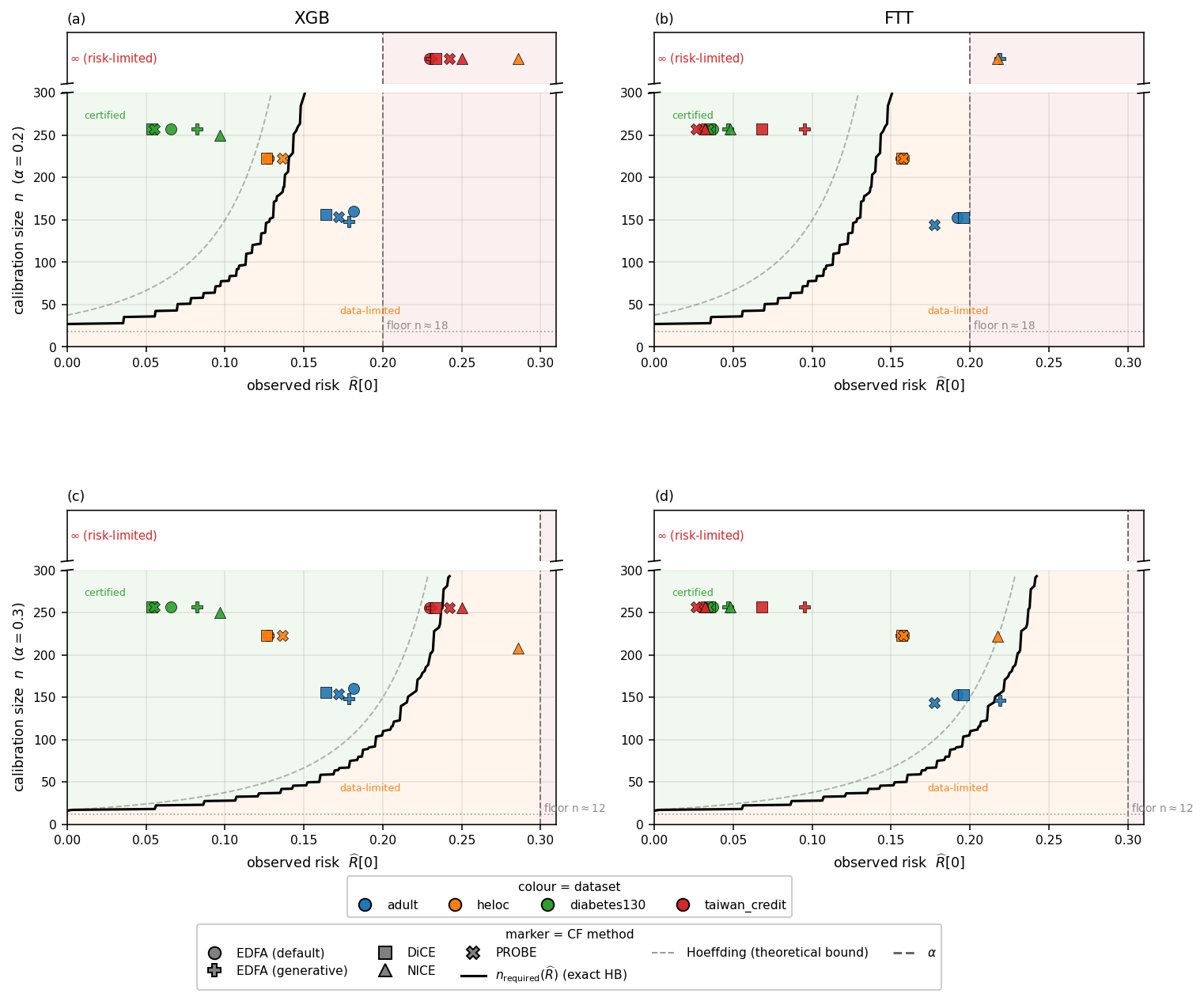}
    \caption{This shows relationship of calibration set size and the observed risks, in the selected datasets. Similar to figure~\ref{fig:g1} and~\ref{fig:g2}, different CF search methods are used. The vertical dotted line (floor) represents a minimum amount of calibration data required. The curve in the solid black line represent the empirical lower bound for the required data (from exact RCPS-HB), whereas the curve in dotted line is the Hoeffding derived theoretical curve (a stricter bound). The risk-limited (red) zone shows model-CF-dataset combinations which cannot provide the requested guarantee, even more data is collected and used. The data-limited (yellow) zone represents the model-CF combinations which are currently unable to provide the guarantees, but is able to do so with the help of more calibration data. The amount of additional data required is theoretically bounded by eq~\ref{eq:bound1}. The certified (green) zone contains the instances where the guarantee was provided.\\
    As the $\alpha$ level is relaxed from 0.2 to 0.3, more model-CF combinations qualify to provide guarantees with the current amount of calibration data. FTT model is able to provide guarantees better than the XGB model, in general.}
    \label{fig:d1}
\end{figure}

This section presents the validation guarantee results for the RCPS-HB method applied to explanations obtained using the proposed EDFA framework. 
These validation guarantees translate into user requirements as follows. Suppose a user receives a denied prediction and a recourse recommendation at an early acquisition step, having supplied only a subset of their features. The certified stopping rule then offers the following contract: among users who act at the first step where the model's uncertainty falls below the certified threshold $\hat\tau$ at most an $\alpha$ fraction will find that either the prediction itself would have changed or the recommended recourse would no longer be valid had all remaining features been acquired. This holds with confidence of $1 - \delta$ over the draw of the calibration data. At $\alpha =$ 0.2 and $\delta =$ 0.05, for instance, at least 80\% of early-stopping users receive recourse that survives full information, certified at the 95\% confidence level.
The computation of validation guarantees requires a separate, untouched calibration set. The CF search method used in the EDFA algorithm is varied in these experiments.  Figures~\ref{fig:g1} and \ref{fig:g2} show the average validation guarantee provided by each CF search method with the proposed EDFA, at different steps of the AFA process. The steps correspond to average costs incurred at each step denoted by the values on the x-axis of each subfigure. It can be observed that CF search algorithms, including EDFA can indeed generate recourse with the expected validation guarantee $\alpha=0.2$ or $0.3$) without having to spend the full budget of feature acquisition. For instance, CF search methods generate such recourse for the diabetes dataset  (figures~\ref{fig:g1}(c) and (g)). For the same dataset, the PROBE CF method provides more lower cost recourse options with such guarantees. It is also possible that CF search methods have to spend the full budget in order to generate the guarantees. Certain methods fail to do so even after spending the full budget. For instance, the NICE method falls behind for both $\alpha=0.2$ and $0.3$ on the HELOC dataset (figure~\ref{fig:g1}(b) and (f)).\\
The second set of figures (figure~\ref{fig:d1}) shows the data limitations of the calibration sets used to compute validation guarantees. As it was shown earlier, the theoretical lower bound of the required calibration set size is derived using the equation~\ref{eq:hb-lower}. This lower bound is stricter as mentioned in the theoretical background section~\ref{sec:theory}. Hence, the empirically calculated curve is used which approximates the shape of the theoretical curve (eq.~\ref{eq:hb-lower}). In each subfigure, the yellow (data-limited) and red (risk-limited) zones indicate the instances where the current amount of calibration data is insufficient to provide a validation guarantee at a given $\alpha$ level. The red zone specifically denotes that any additional data amount cannot yield the guarantees, because the observed risk itself is higher than the given $\alpha$ level. One of the prominent patterns are that the Adult Income dataset can be regularly found in the data-limited zone for the recourse derived for the XGB model (figure~\ref{fig:d1}(a),(c)). Another pattern is that the NICE CF search method on HELOC dataset also follows a similar pattern (figure~\ref{fig:d1}(a),(b),(c)).

\subsection{Ablation studies}
\begin{figure}[htbp]
     \begin{subfigure}{\linewidth}
        \centering
        \includegraphics[width=\linewidth]{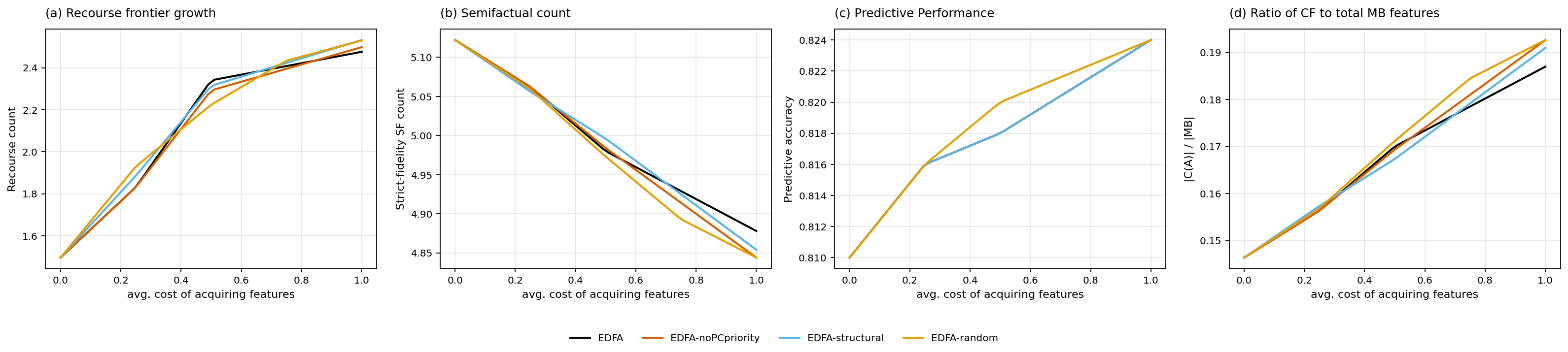}
        \caption{Adult Income - FTT}
        \label{fig:adult22abl}
    \end{subfigure}

    \begin{subfigure}{\linewidth}
        \centering
        \includegraphics[width=\linewidth]{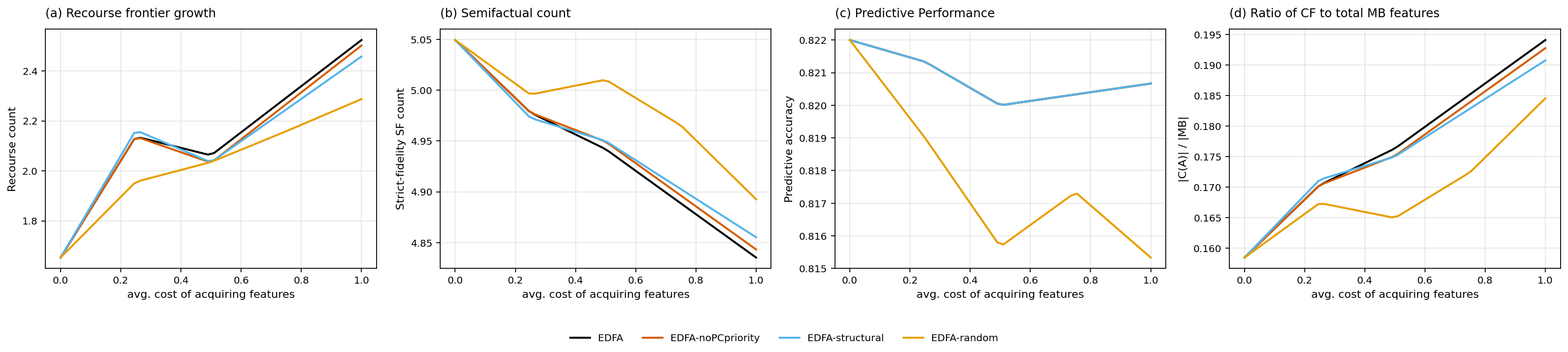}
        \caption{Adult Income - XGB}
        \label{fig:adult21abl}
    \end{subfigure}

    \caption{Ablation experiment results obtained for the Adult Income dataset on (i) FTT model, (ii) XGB model. The EDFA-random variant shows less recourse options and worse accuracy performance than other variants of EDFA, specifically for the XGB model. However, the variants are comparatively stable for the FTT model.}
    \label{fig:abladult}
\end{figure}

\begin{figure}[htbp]
    \begin{subfigure}{\linewidth}
        \centering
        \includegraphics[width=\linewidth]{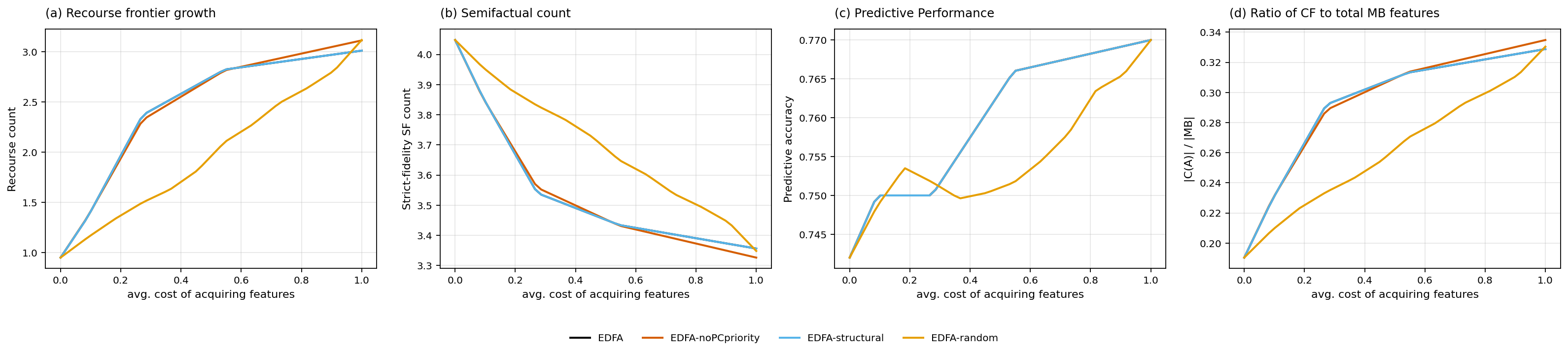}
        \caption{HELOC - FTT}
        \label{fig:heloc22abl}
    \end{subfigure}

     \begin{subfigure}{\linewidth}
        \centering
        \includegraphics[width=\linewidth]{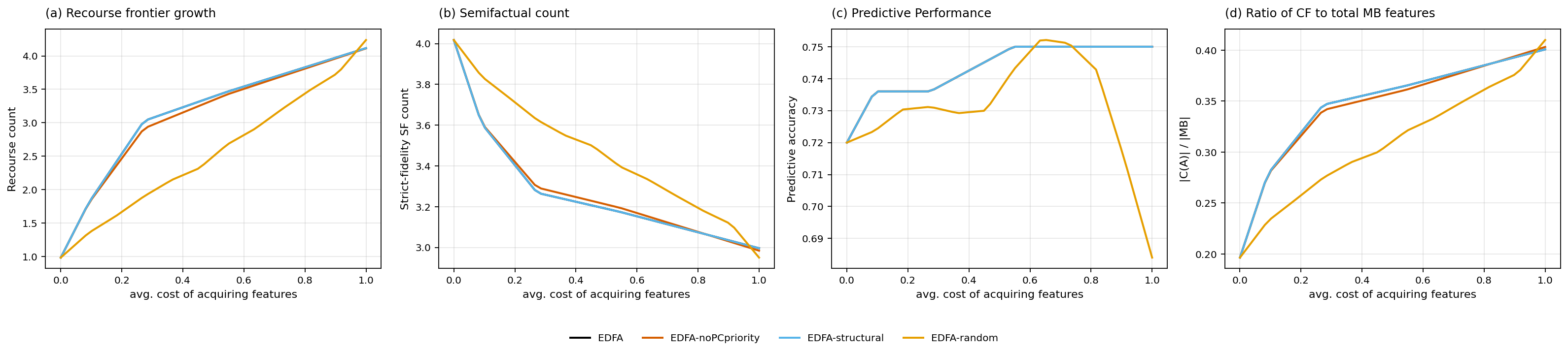}
        \caption{HELOC - XGB}
        \label{fig:heloc21abl}
    \end{subfigure}
    \caption{Ablation experiment results obtained for the HELOC dataset on (i) FTT model, (ii) XGB model. The EDFA-random variant provides low recourse count and unstable accuracy performance.}
     \label{fig:ablheloc}
\end{figure}

\begin{figure}[htbp]
 \begin{subfigure}{\linewidth}
        \centering
        \includegraphics[width=\linewidth]{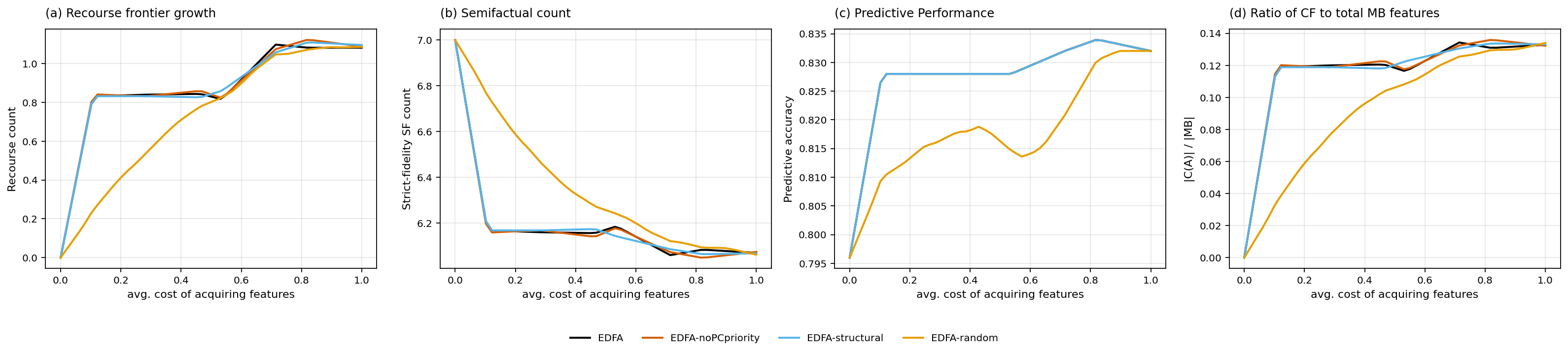}
        \caption{Taiwan Credit - FTT}
        \label{fig:taiwan22abl}
    \end{subfigure}
 \begin{subfigure}{\linewidth}
        \centering
        \includegraphics[width=\linewidth]{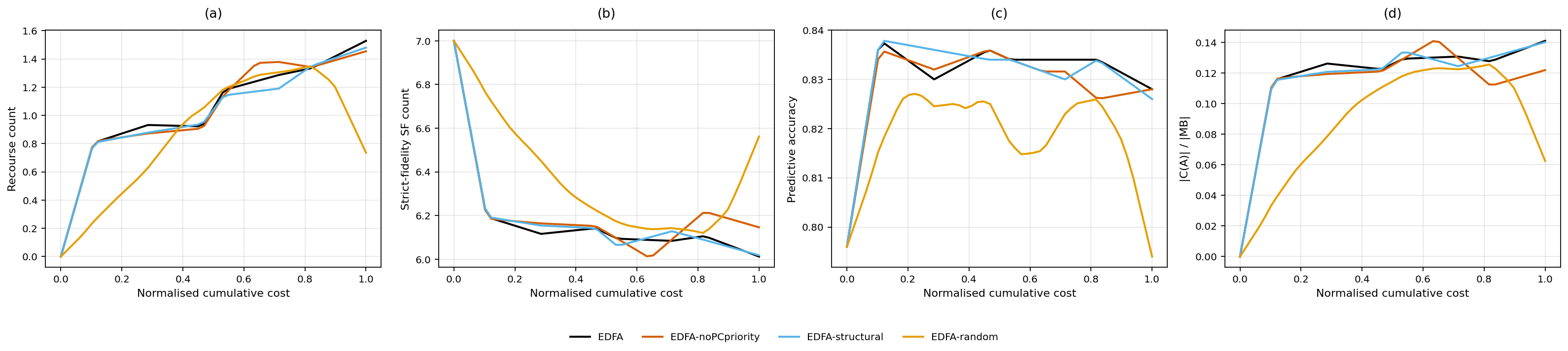}
        \caption{Taiwan Credit - XGB}
        \label{fig:taiwan21abl}
    \end{subfigure}

    \caption{Ablation experiment results obtained for the Taiwan Credit dataset on (i) FTT model, (ii) XGB model. The degraded performance of EDFA-random is prominent here.}
     \label{fig:abltaiwan}
\end{figure}

\begin{figure}[htbp]
    \begin{subfigure}{\linewidth}
        \centering
        \includegraphics[width=\linewidth]{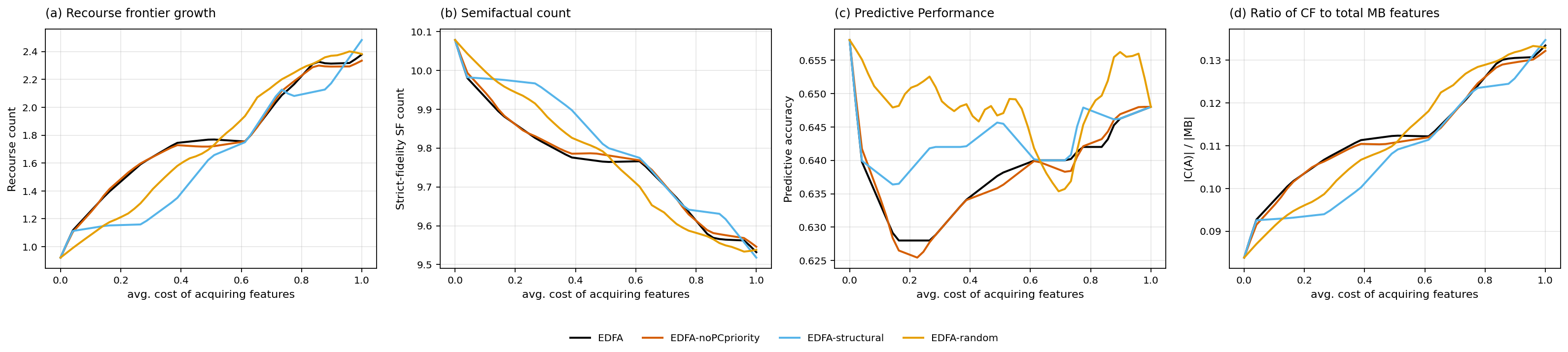}
        \caption{Diabetes130 - FTT}
        \label{fig:diabetes22abl}
    \end{subfigure}

     \begin{subfigure}{\linewidth}
        \centering
        \includegraphics[width=\linewidth]{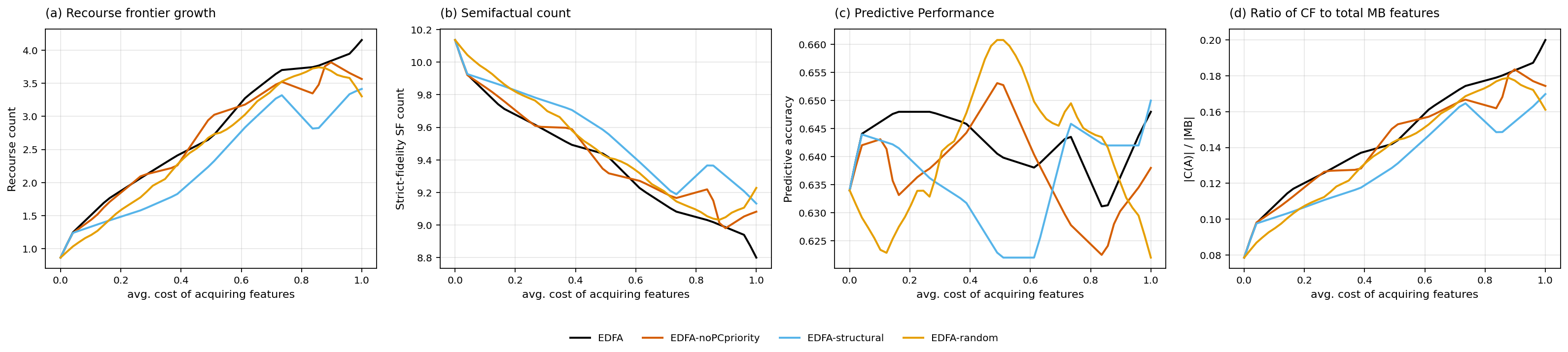}
        \caption{Diabetes130 - XGB}
        \label{fig:diabetes21abl}
    \end{subfigure}

    \caption{Ablation experiment results obtained for the Diabetes dataset on (i) FTT model, (ii) XGB model. EDFA-random shows performance on par with other variants for this dataset.}
     \label{fig:abldiabetes}
\end{figure}

The ablation study is focused on how each core component of the EDFA algorithm~\ref{alg:edfa} is crucial for its performance (figures~\ref{fig:abladult}-\ref{fig:abldiabetes}). Three variants of the EDFA algorithm 
are tested; EDFA-random, EDFA-noPCpriority and EDFA-structural. EAMB was used as the underlying MB-discovery method for the ablation experiments. EDFA-random variant was also used as a baseline, where the features are picked in a uniform random manner, from the MB. EDFA-noPCpriority removes the requirement to acquire $PC$ nodes first from an MB-unit (line 13 of algorithm~\ref{alg:edfa}). EDFA-structural prioritize $PC$ members but do not use CMI/cost (line 16 of algorithm~\ref{alg:edfa}), instead only use 1/cost.\\
For majority of the datasets, all the three variants do not show significant deviations from the vanilla EDFA algorithm. However, as the dataset feature count goes up, EDFA-random variant shows degraded performance (lower recourse count, and lower accuracy) at certain stages of the AFA, compared to the other EDFA variants in Taiwan Credit and HELOC dataset experiments, for both XGB and FTT models (figures~\ref{fig:ablheloc}-\ref{fig:abltaiwan}). However, the it shows slightly better performance in terms of accuracy for the diabetes dataset.\\
These results align closely with Proposition 2. First, EDFA-noPCpriority (which removes the structural PC-first enforcement (line 13 of Algorithm 1) but retains CMI/cost ranking), tracks the vanilla EDFA trajectories almost exactly on all datasets (figures~\ref{fig:abladult}-\ref{fig:abltaiwan}). This is the behavior Proposition 2 predicts; since the structural ordering and the CMI-greedy ordering coincide within an MB unit, removing the explicit enforcement should change little whenever per-instance CMI estimates are reliable. The enforcement in Algorithm 1 thus guards against CMI estimation noise rather than a distinct policy, and the ablation confirms the two implementations are empirically interchangeable in our settings. Second, EDFA-structural, which replaces the CMI/cost spouse ranking with 1/cost, likewise shows no significant deviation, indicating that the unit-level acquisition ordering, rather than the precise information-gain magnitudes, drives EDFA's recourse growth. Third, the contrast with EDFA-random delimits the scope of these equivalences: as the feature count grows(figures~\ref{fig:abltaiwan},\ref{fig:ablheloc}), random selection within the MB degrades in both recourse count and accuracy. MB membership alone is therefore not sufficient; the MB-unit structure exploited by Propositions 1–2 becomes increasingly important as the blanket grows. The Diabetes dataset, where EDFA-random remains competitive (figure~\ref{fig:abldiabetes}), is consistent with this reading, as its large MB (11 features) contains many counterfactually active members, so even undirected selection within the blanket would have a higher chance to encounter them.

\section{Discussion}
\label{sec:discussion}
\begin{figure}[htbp]
    \centering
    \begin{subfigure}[b]{0.22\textwidth}
        \centering
        \includegraphics[width=\linewidth]{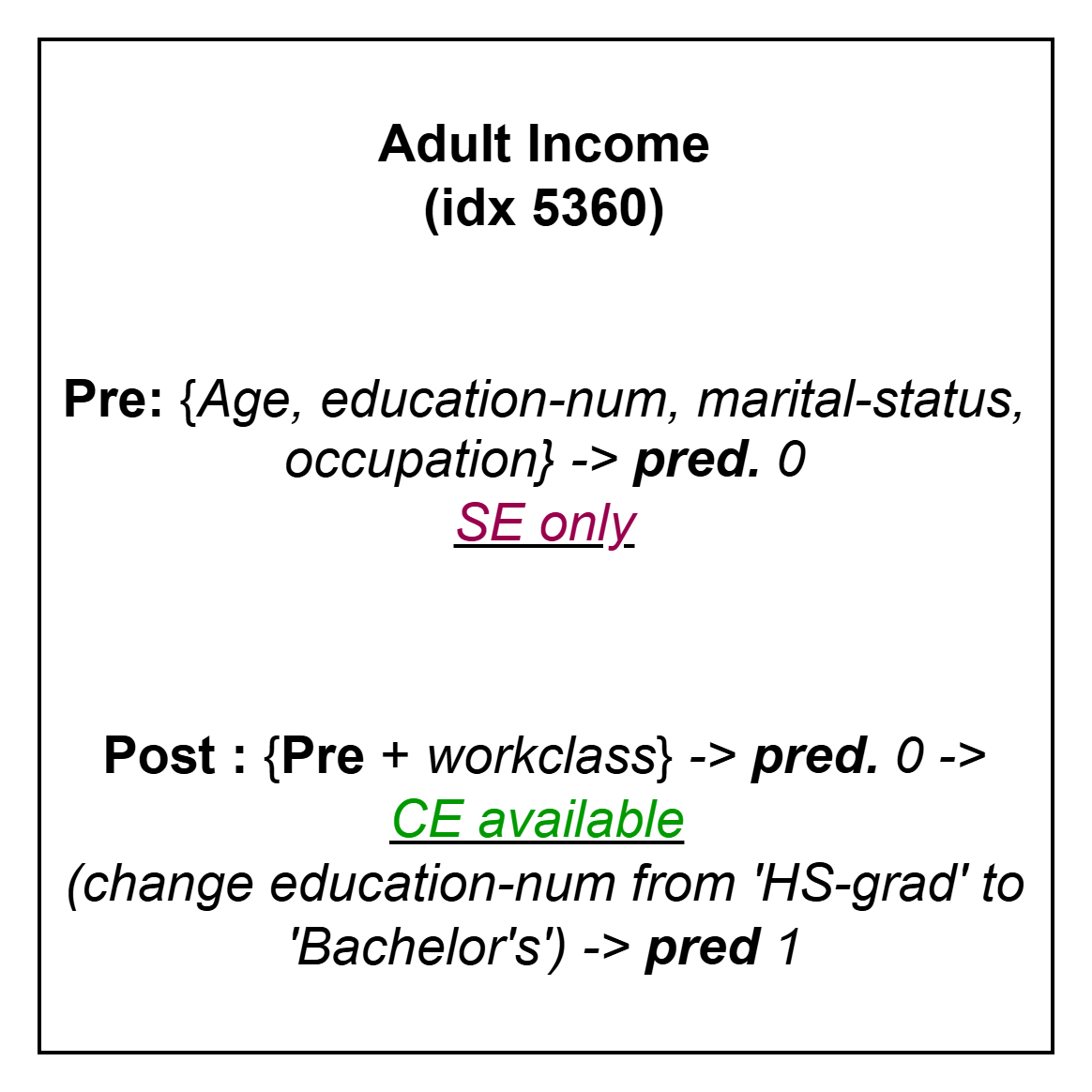}
        \subcaption{}
        \label{fig:transition1}
    \end{subfigure}%
    \begin{subfigure}[b]{0.22\textwidth}
        \centering
        \includegraphics[width=\linewidth]{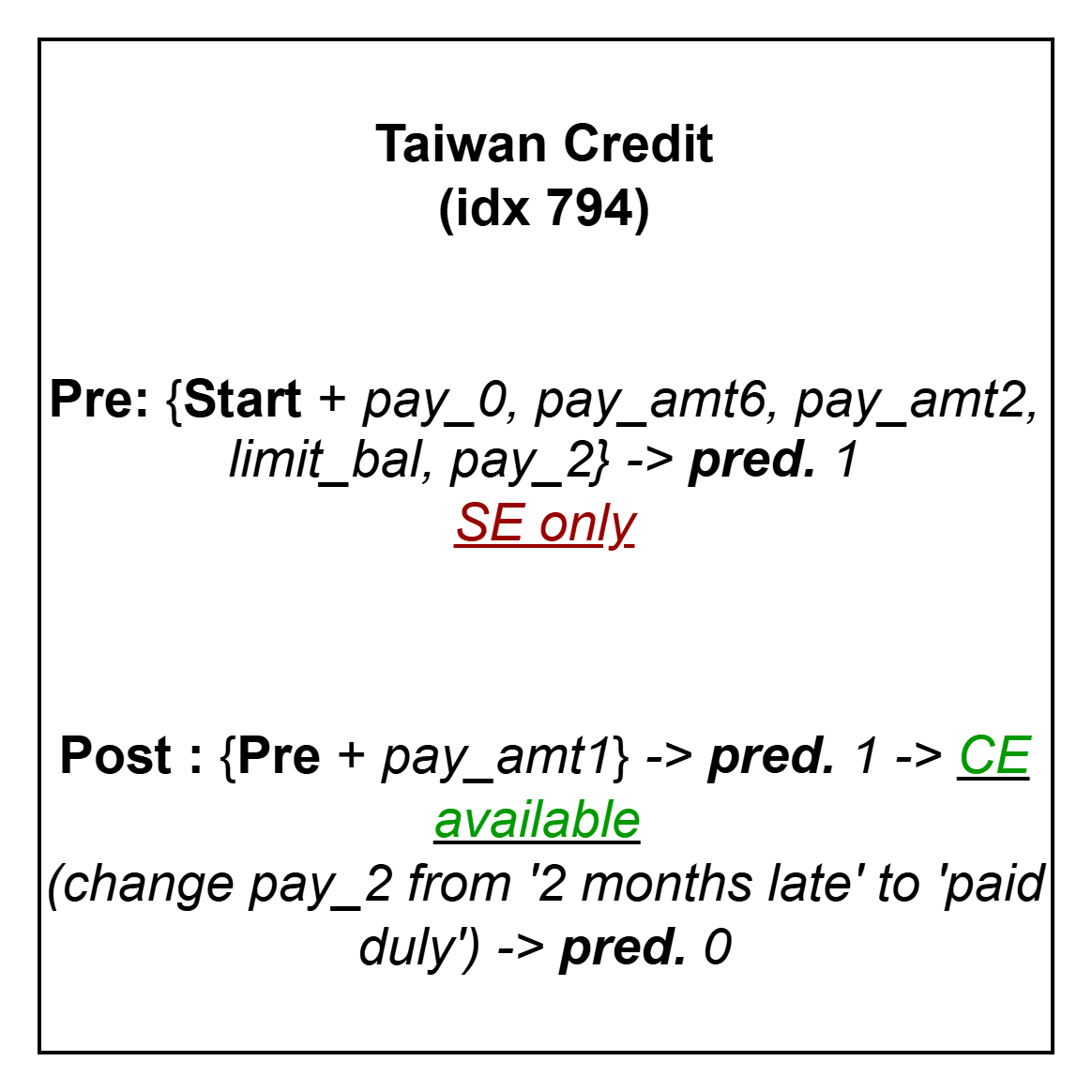}
        \subcaption{}
        \label{fig:transition2}
    \end{subfigure}
    \begin{subfigure}[b]{0.22\textwidth}
        \centering
        \includegraphics[width=\linewidth]{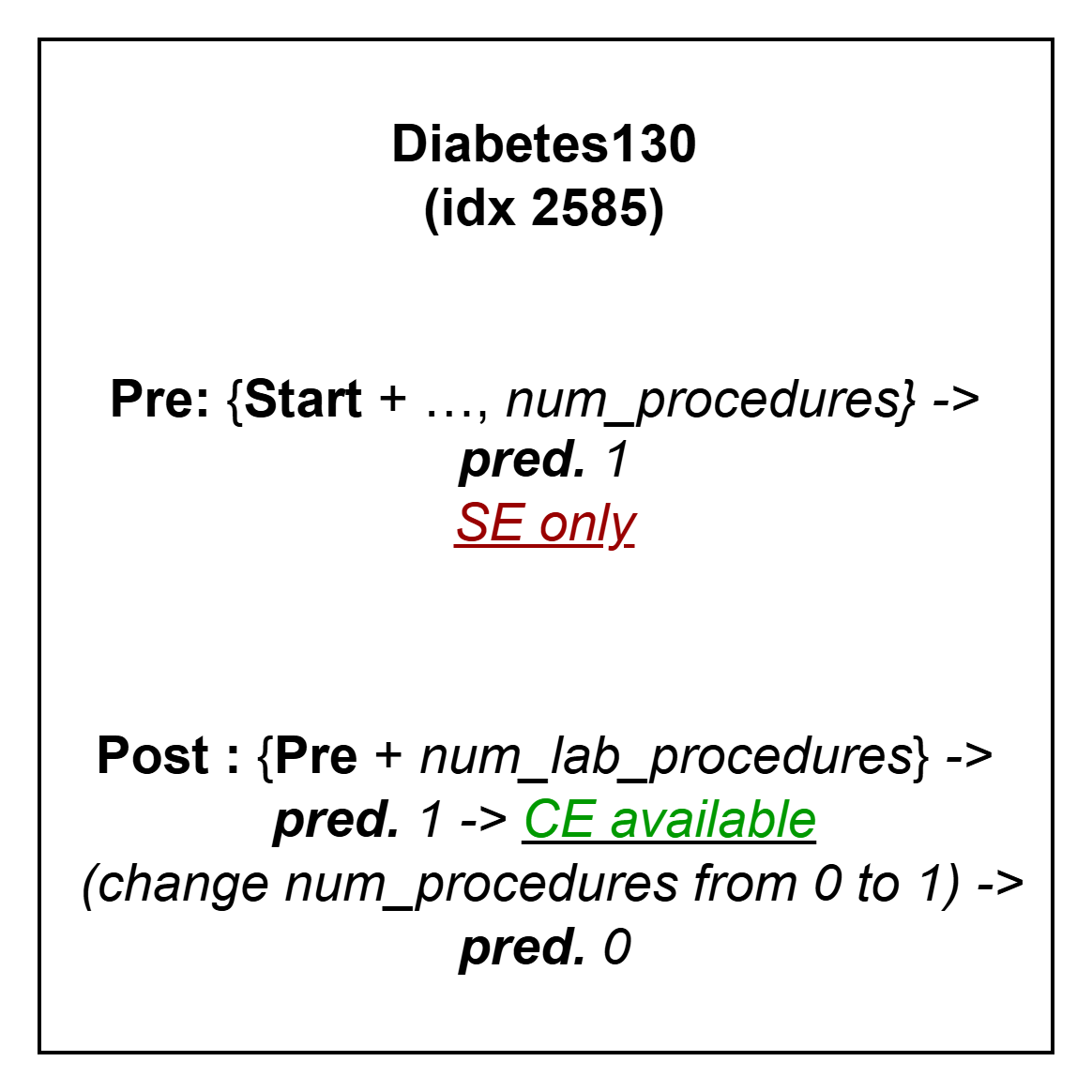}
        \subcaption{}
        \label{fig:transition3}
    \end{subfigure}
    \begin{subfigure}[b]{0.22\textwidth}
        \centering
        \includegraphics[width=\linewidth]{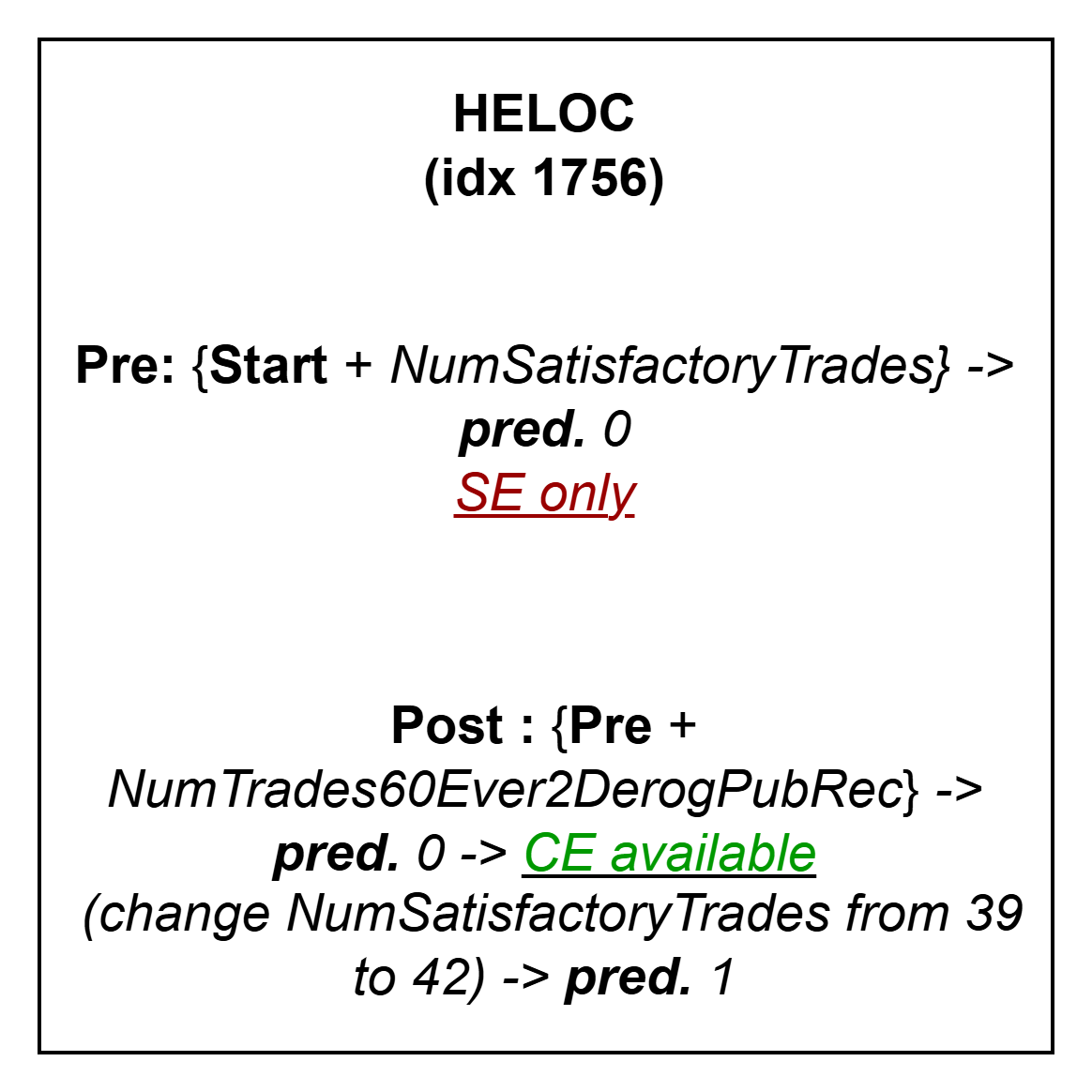}
        \subcaption{}
        \label{fig:transition4}
    \end{subfigure}
    
    \caption{The examples extracted from the datsets, demonstrating how the SF to CF transition occurs with feature acquisition in EDFA. In each instance, \textbf{Pre} denotes the features available at the present state, and only SFs are available for each instance. \textbf{Post} denotes the state after acquiring a feature, which enables CFs as well.\\
    For instance, figure (i) shows the transition of a test instance extracted from the Adult Income dataset. The starting features (\textbf{Pre}) are unable to provide any recourse and currently predicted as label 0 (income <50K). The EDFA algorithm picks `workclass' feature (\textbf{Post}), which still classifies the instance as label 1, but also create the potential for recourse (by perturbing the previous `education-num' feature)}
    \label{fig:transitions}
\end{figure}

\begin{figure}[htbp]
    \centering
    \begin{subfigure}[b]{0.3\textwidth}
        \centering
        \includegraphics[width=\linewidth]{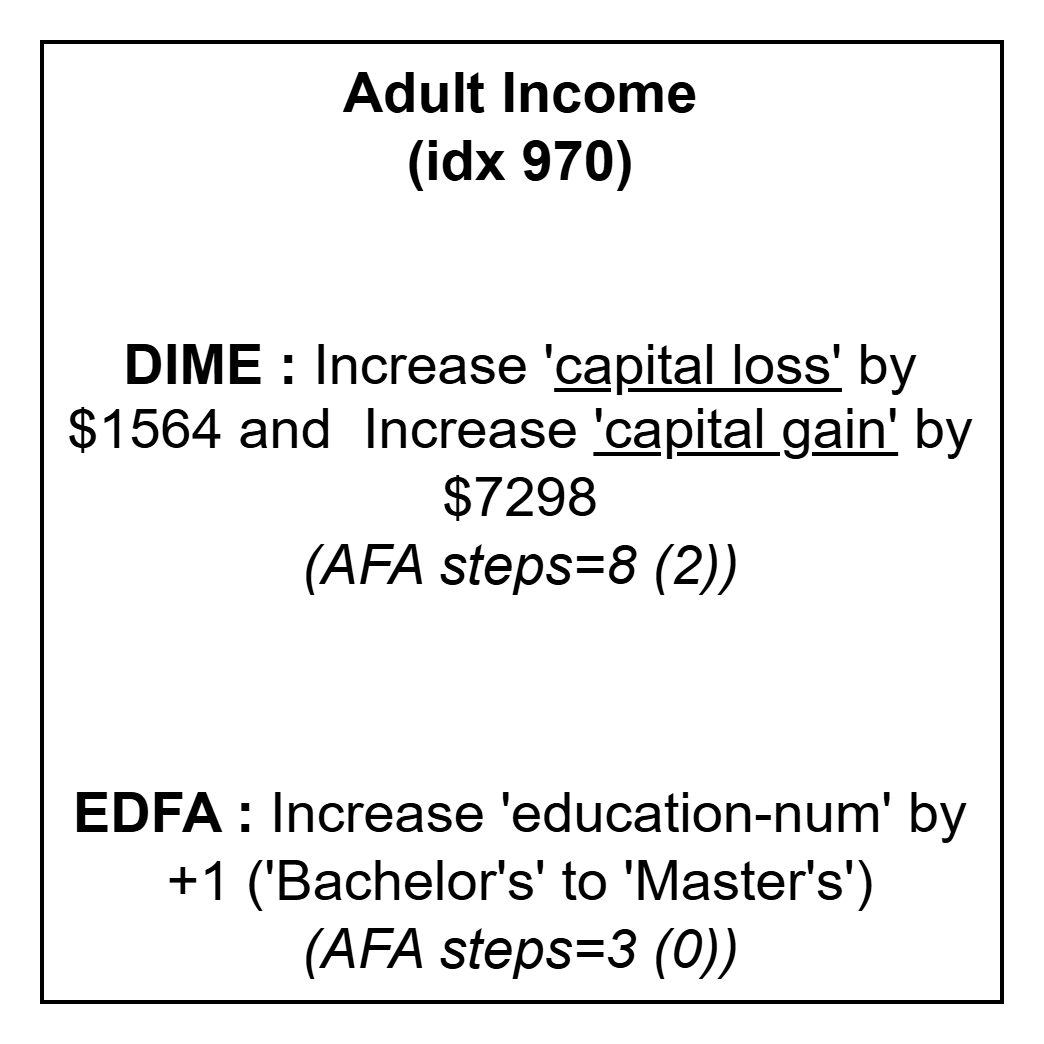}
        \subcaption{}
        \label{fig:mbexample1}
    \end{subfigure}%
    \hfill %
    \begin{subfigure}[b]{0.3\textwidth}
        \centering
        \includegraphics[width=\linewidth]{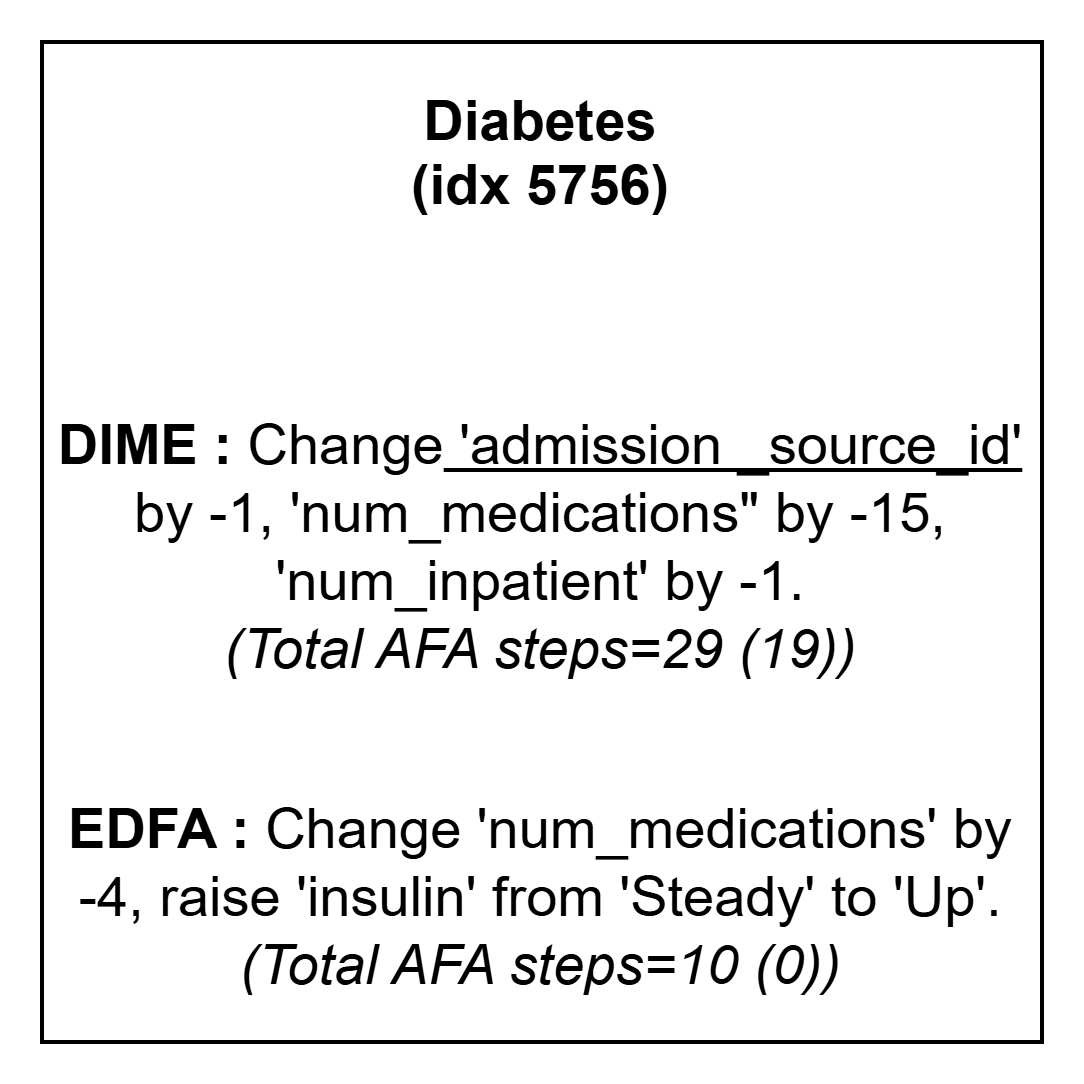}
        \subcaption{}
        \label{fig:mbexample2}
    \end{subfigure}
    \hfill%
    \begin{subfigure}[b]{0.3\textwidth}
        \centering
        \includegraphics[width=\linewidth]{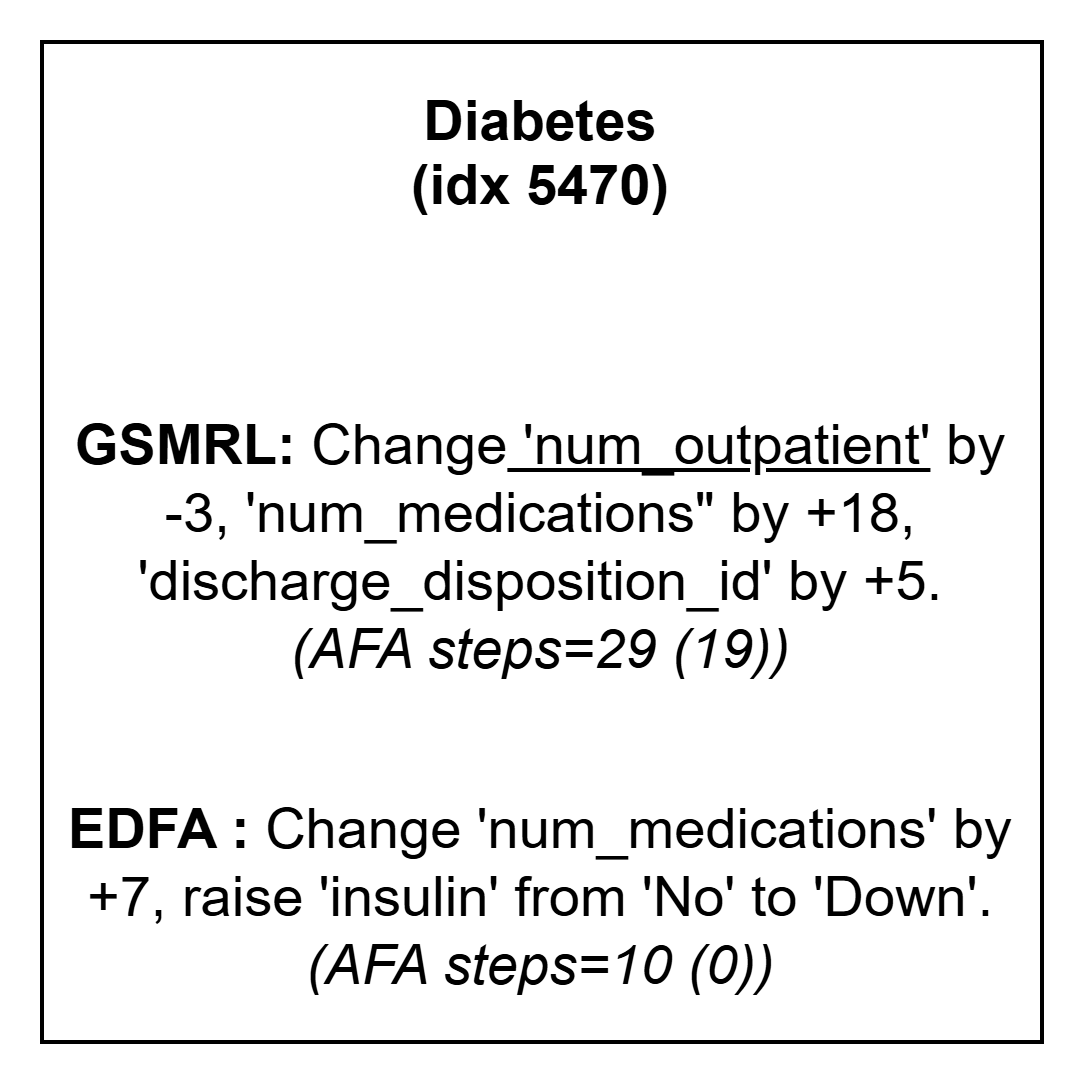}
        \subcaption{}
        \label{fig:mbexample3}
    \end{subfigure}
    
    \caption{These figures show a few sample instances extracted from the recourse options generated by AFA baseline methods and the proposed EDFA. Each recourse instance is accompanied by the `AFA steps' count incurred by the baseline and the EDFA methods. The value next to the AFA steps and within brackets are the number of non-MB features acquired. For instance, in Figure (i), the recourse options provided by the baseline AFA method DIME suggest actions that are less practical, increasing capital loss and the gain simultaneously. EDFA proposes a more plausible action. DIME takes 8 AFA steps and acquires 2 non-MB features along the way, whereas EDFA spends only 3 AFA steps and does not acquire any non-MB feature.}
    \label{fig:examples}
\end{figure}

DIME, EDDI and GSMRL provide higher counts of recourse options, with an increasing reliance on non-MB features as AFA progresses. This can be attributed to that these methods do not have access to the discovered MB nor pick features considering the MB structure. This proves that the proposed EDFA algorithm successfully picks features that are structurally connected and within the MB. The implications of this difference is that the EDFA algorithm picks more `decision-relevant' features, that are also faithful to the MB. The baseline AFA methods may select decision-irrelevant features that are non-actionable in some cases. For instance, we identify several such cases from the experiments, which are shown in figures~\ref{fig:examples}. For instance, the first example shows two recourses provided for the same test instance, one by perturbing the features acquired with the DIME method and the second provided by the EDFA method (figure~\ref{fig:mbexample1}). The DIME AFA method picks non-MB features on the Adult Income dataset (figure~\ref{fig:adult21}), and provides a recourse option with two conflicting actions to increase the capital gain as well as the capital loss.\\
The baseline comparison figures confirm that the change of the MB discovery algorithm does not change the major observations pertaining to the EDFA and other AFA methods. In fact, the change to the MB discovery algorithm does not significantly alter the recourse options. This confirms that the EDFA is robust to the MB discovery method as well. The core reason is that the MB discovery methods recover the the same set of MB features each time. However, the only difference comes from the structure of the learned MB. This can be further seen in the extracted transitions shown in figure~\ref{fig:transitions}. The first example shown for the Adult Income dataset was derived using HITON-MB as the MB discovery method, whereas the other two examples used EAMB. It turned out that EAMB did not recover the MB-unit structure exactly similar to HITON-MB, for the Adult Income dataset. Hence, we acknowledge this as a possible limitation, which can be overcome by manually verifying if the MB-units recovered by each MB discovery method align with decision-relevant or actionable features.

It can be observed that the EDFA method is on par with other AFA methods in terms of predictive performance, as shown in the subfigures that show the accuracy throughout AFA for each method in figure~\ref{fig:res1}-\ref{fig:res2}. This performance is achieved even though EDFA does not rely solely on prediction gain in its algorithm, unlike other AFA methods. This also implicitly indicates that the features selected by EDFA are relevant to predictive performance. Notably, EDDI method lags behind all other AFA methods, which is consistent with the results reported in some previous experiments in the literature~\cite{gadgil2024estimating}. As shown in figures~\ref{fig:gmsc1}, and figures~\ref{fig:gmsc21} and~\ref{fig:gmsc22}, the GMSC dataset shows poor performance in terms of the valid recourse options found by the EDFA method. The sizes of the MBs found for the GMSC dataset are low (only 3 features in the MB), which signals that MBs without sufficient CF features (or MBs that only have semifactual features in them) cannot provide recourse options. In other words, this means that the discovered MBs for the GMSC dataset do not contain label-flipping features, which aligns with the fact that not all features inside an MB would be label flipping (Theorem 1). Taken together, the ablations indicate that EDFA method's advantage over baseline AFA methods stems from where it looks (the MB) and in what order (PC-first, per Proposition 2).\\
The validation guarantees provide a theoretically grounded way to quantify whether a user can rely on a recourse (and its associated original prediction) when not all features are available (e.g., due to cost or time constraints). The results confirm that it is possible to obtain such a guarantee at a given risk level without spending the full budget or acquiring all the features. However, the availability of guarantees largely depends on the dataset and is also constrained by its size. Furthermore, the choice of CF search method does not seem to affect the validation guarantees significantly, except that the PROBE CF method achieves them at a lower cost than other methods on the diabetes dataset. Nonetheless, the risk-limited scenarios suggest changing the CF search method or the AFA method itself. For instance, the EDFA-generative CF search method is unable to provide validation guarantees for the Adult Income dataset with the FTT model, even with additional calibration data (risk-limited) (figure~\ref{fig:d1} (b)). However, choosing EDFA-default, PROBE, or DiCE as the CF search method makes it solvable by collecting more data. 

One limitation of the proposed EDFA algorithm is that it may provide valid recourse (i.e., it flips the prediction) but not truly actionable. The current setup assumes (Assumption 5) that the features acquired within the MB are actionable, which may not hold in certain cases. EDFA can yield such non-actionable recourse for the following reason. As the feature transition can happen in two ways (Definition 9), either spouse ($Sp$) features can be made actionable (by acquiring the corresponding $Ch$ node and completing a v-structure (Proposition 1)) or by completing the MB unit and activating PC members. Among these, $Sp$ features in particular can be exogenous context variables (e.g., length of stay in the hospital) that are not actionable. This is also why the example transitions shown (figures~\ref{fig:transitions}) are MB-unit completion examples. Incorporating directionality filtering (or directional feasibility) over $C(A)$ to gauge true actionability remains future work.
The validation guarantees have two limitations: first, the guarantee is marginal, not conditional. This means that the guarantee signals the population failure rate but not the the risk faced by an individual user. Second, the guarantee concerns about the stability of the prediction and the recourse under further acquisitions, but its true actionability still depends on meaning of the feature in the real world. This should be addressed as mentioned above. However, under these limitations the guarantees still answer the question that motivates early stopping in the first place; whether a user can act now, at lower cost, without the decision being changed later once the remaining information arrives.

Strategic Manipulation of CFs (gaming the systems) is a known challenge with recourse, in which users exploit the CFs provided to them to gain an unfair advantage and fool the system~\cite{venkatasubramanian_philosophical_2020, slack_counterfactual_2021}. One possible scenario is that, in certain applications such as credit approvals, users may find feature combinations that assist such manipulations. Baseline AFA methods may allow such manipulations because they pick non-MB features that are correlated with the prediction but not decision-relevant. EDFA, by contrast, limits manipulability by restricting perturbable features to the MB. While we do not evaluate this quantitatively in this work, it is worth focusing on in the future. Another useful inspection would involve users specifying the features or settings that are actionable for them. Our current work assumes a global setting in which all users have the same starting and actionable features.  

\section{Conclusion}
\label{sec:conclusion}
This work presented the first exploration of the link between algorithmic recourse and active feature acquisition methods. In particular, we unified three types of algorithmic recourse using Markov Blanket theory and introduced a novel explanation-driven feature acquisition method (EDFA) to explore this relationship. The empirical experiments demonstrated that the proposed method can provide mode `decision-relevant' recourse while maintaining predictive performance. Furthermore, we proposed an idea and computational method for obtaining validation guarantees for early recourse. Future work can extend this work to additional application scenarios and datasets, particularly multimodal datasets useful in applications such as medical diagnosis. Moreover, studying other properties of the CFs, such as their distance (which measures the effort to realize the actions), is also of interest. Utilizing causal frameworks beyond Markov Blankets is another interesting avenue, which could ensure actionability by incorporating a more rigorous framework that considers directionality between features as well. 

\section*{Acknowledgments}
-

\bibliographystyle{unsrt}  
\bibliography{references}  

\appendix
\begin{landscape}
\begin{table}[h]
\centering
\caption{Feature cost categories per dataset, ranging from free to sensitive (banned) features.}
\label{tab:datasets2}
\begin{tabularx}{1.6\textwidth}{|X|X|X|X|X|X|}
     \hline
     Dataset & Starting (Free) & Cheap (1) & Moderate (2) & Expensive (5) & Sensitive (Banned)\\
     \hline
     German Credit &age, credit amount, purpose &employment, other parties, residence since, housing, job, num dependents, own telephone &checking status, duration, credit history, savings status, installment commitment, property magnitude, other payment plans,
  existing credits&personal status, foreign worker&personal status, foreign worker, age\\
     Adult Income & age, education-num, marital-status, occupation & workclass, education, relationship, native-country & capital-gain, capital-loss, hours-per-week &race, sex & race, sex, native-country, relationship, marital-status\\
     HELOC & ExternalRiskEst., AverageMInFile, NumTotalTrades& MSinceOldestTr.Op., MSinceMostRecentTr.Op, NumSatisfactoryTrades& Perc.TradesNeverDelq, NumTradesOpeninLast12M, Perc.InstallTrades, NetFrac.RevolvingBurd., NetFrac.InstallBurd.,
  NumRev.TradesWBal., NumInstallTradesWBal., NumBank2NatlTradesWHighUtil., Perc.TradesWBal.,
  MSinceMostRecentInqexcl7days, NumInqLast6M, NumInqLast6Mexcl7days& NumTrades60Ever2DerogPubRec, NumTrades90Ever2DerogPubRec, MSinceMostRecentDelq, MaxDelq2PublicRecLast12M& N/A\\
     Taiwan Credit &age, sex, education, marriage & Same as the starting set& limit bal , pay 0, bill amt1, pay amt1 (recent repayments) & pay 2…pay 6, bill amt2…bill amt6, pay amt2…pay amt6 — 15 (older repayment/bill/payment history)&sex, age, marriage, education \\
     ACS Income & AGEP, SEX, RAC1P&MAR, RELP & COW, SCHL, WKHP · (3, detailed coded) OCCP, POBP& N/A & SEX, RAC1P, AGEP, MAR, RELP\\
     GMSC&age, NumberOfDependents& Same as the starting set&RevolvingUtilizationOfUnsecuredLines, DebtRatio, MonthlyIncome, NumberOfOpenCreditLinesAndLoans, NumberRealEstateLoansOrLines&NumberOfTime30-59DaysPastDueNotWorse, NumberOfTime60-89DaysPastDueNotWorse, NumberOfTimes90DaysLate&age\\
     Diabetes130 &race, gender, age, time in hospital, number outpatient, number emergency, number inpatient & admission type id, discharge disposition id, admission source id& num lab procedures, num procedures, num medications, number diagnoses, change, diabetesMed + 21 individual drug columns
  (metformin, glipizide, glyburide, insulin, … incl. combo meds)& max glu serum, A1Cresult& race, gender, age\\
     \hline
\end{tabularx}
\end{table}
\end{landscape}

\section{Proofs of the Theorem 1, Propositions 1 and 2}
\label{sec:proof}

Throughout, $G=(V,E)$ is the DAG over $V=\{Y,X_1,\dots,X_n\}$ with joint distribution $P$, and $MB(Y)=Pa(Y) \bigcup Ch(Y) \bigcup Sp(Y)$ as in Definition~1. At an acquisition state with acquired set $A$ and observed values $x_A$, the model prediction is $y^{*}=f(x_A)$. By the aligned-model assumption (Assumption~3) the predictor is trained on the Markov-blanket features only; on its inputs it realises the Bayes rule $f(x_A)=\argmax_{y} P(Y=y \mid X_{A \cap MB(Y)}=x_{A \cap MB(Y)})$ and is invariant to every coordinate outside $MB(Y)$. We use three standard consequences of the assumptions:
\begin{itemize}
    \item[(T1)] \textbf{(Soundness of $d$-separation, from Assumption~1.)} If a set $X_3$ $d$-separates $X_1$ from $X_2$ in $G$, then $X_1$ and $X_2$ are conditionally independent given $X_3$ under $P$.
    \item[(T2)] \textbf{(Faithfulness, Assumption~2.)} If a path between $X_1$ and $X_2$ is active (d-connecting) given $X_3$, then $X_1$ and $X_2$ are statistically dependent given $X_3$. This is the contrapositive of Assumption~2.
    \item [(T3)]\textbf{(Markov-blanket shielding)}~\cite{kollerProbabilistic2009}Under Assumption~1, $X_{MB(Y)}$ $d$-separates $Y$ from every variable outside the blanket; hence $Y$ is conditionally independent of $X_{V \backslash (MB(Y) \cup \{Y\})}$ given $X_{MB(Y)}$, and
\[
   P\!\left(Y \mid X_{MB(Y)}\right)=P\!\left(Y \mid X_{V \backslash \{Y\}}\right).
\]
\end{itemize}




\subsection{Proof of Theorem~1}

\textbf{(i) Sufficiency: $\beta(MB(Y),x,y^{*})=1$ for all $x$.}
Write $S=MB(Y)$ and $\bar S = V \backslash (MB(Y) \cup \{Y\}) = \mathcal{A}_{\text{alt}}$, so that $y^{*}=f(x_S,x_{\bar S})$. Lemma~A.1 shows that aligning the predictor to the blanket is lossless: $P(Y \mid X_{MB(Y)})=P(Y \mid X_{V \backslash \{Y\}})$, so the blanket-restricted Bayes rule is also Bayes-optimal given all features. By Assumption~3 the predictor is trained on $MB(Y)$ only and is therefore invariant to $\bar S$: for every replacement $x'_{\bar S}$,
\[
   f(x_S,x'_{\bar S})=f(x_S,x_{\bar S})=y^{*}.
\]
The indicator inside the observational-sufficiency functional $\beta$ is then identically $1$, giving $\beta(MB(Y),x,y^{*})=1$ for all $x$. \hfill$\square$

\textbf{(iii) Alterfactual correctness: $\alpha(\{X_i\},x,y^{*})=0$ for all $x$, whenever $X_i \in \mathcal{A}_{\text{alt}}$.}
If $X_i \in \mathcal{A}_{\text{alt}} = V \backslash \{Y\} \backslash MB(Y)$ then $X_i \notin MB(Y)$, so by Assumption~3 the predictor does not depend on $X_i$. For any replacement $x'_i$ we have $f(x'_i,x_{-i})=f(x_i,x_{-i})=y^{*}$, so the indicator inside the observational-necessity functional $\alpha$ is identically $0$ and $\alpha(\{X_i\},x,y^{*})=0$ for all $x$. \hfill$\square$

\textbf{(ii) Individual necessity.}
Fix $X_i \in MB(Y)$.

\textit{Step 1 (probabilistic relevance; Assumptions~1--2).} We first show that $X_i$ remains informative about $Y$ given the rest of the blanket, i.e.\ $I\!\left(Y;X_i \mid X_{MB(Y) \backslash \{X_i\}}\right)>0$. Condition on $X_{MB(Y) \backslash \{X_i\}}$ and distinguish the membership of $X_i$:
\begin{itemize}
   \item if $X_i \in Pa(Y)$ or $X_i \in Ch(Y)$, then $X_i$ is adjacent to $Y$, and adjacent vertices cannot be $d$-separated by any set excluding them, so the connecting edge is an active path;
   \item if $X_i \in Sp(Y)$, there is a child $C \in Ch(Y)$ with $X_i \in Pa(C) \backslash \{Y\}$ forming the collider $Y \rightarrow C \leftarrow X_i$; since $C \in MB(Y) \backslash \{X_i\}$ is in the conditioning set, the collider is open and this v-structure path is active.
\end{itemize}
In either case an active path between $X_i$ and $Y$ exists given $X_{MB(Y) \backslash \{X_i\}}$, so by (T2) the two are dependent and the conditional mutual information is positive. Hence there are a configuration $z$ of $X_{MB(Y) \backslash \{X_i\}}$ and values $a \neq b$ in $\operatorname{supp} P(X_i)$ with $P(Y \mid X_i=a,z) \neq P(Y \mid X_i=b,z)$. Here $\operatorname{supp}$ denotes the probabilistically plausible values of $X_i$ (i.e.- support of $X_i$). \\

\textit{Step 2 (from relevance to a label flip; Assumption~3 and Label flipping Condition).} Under Assumption~3, $f(\cdot)=\argmax_y P(Y=y \mid X_{MB(Y)}=\cdot)$ on the blanket, and the event ``$\alpha(\{X_i\},x,y^{*})>0$ for some $x$'' is \emph{equivalent} to $f$ being non-constant in the $X_i$-coordinate over $\operatorname{supp} P(X_i)$. However,Step~1 does not by itself force this: a probabilistically relevant feature may leave the label unchanged everywhere, since both posteriors in Step~1 can still select the same $\argmax$. The missing ingredient is a non-degeneracy condition, which we state explicitly and which the measured necessity scores corroborate for blanket features.

\textbf{Assumption (Label flipping condition).} \textit{We assume that, a feature $X_i \in MB(Y)$  is CF-relevant (perturbable such that the label flips), if there exist a configuration $z$ of $X_{MB(Y) \backslash \{X_i\}}$ and values $a,b \in \operatorname{supp} P(X_i)$ such that $\argmax_y P(Y=y \mid X_i=a,z) \neq \argmax_y P(Y=y \mid X_i=b,z)$.}

Under this Condition, take the instance $x$ with $x_i=b$ and $x_{MB(Y) \backslash \{X_i\}}=z$ (non-blanket coordinates arbitrary, as $f$ ignores them), so $y^{*}=f(x)=\argmax_y P(Y=y \mid b,z)$. The replacement $a \in \operatorname{supp} P(X_i)$ gives $f(a,x_{-i})=\argmax_y P(Y=y \mid a,z) \neq y^{*}$. The set of replacements whose prediction differs from $y^{*}$ then has positive probability under the sampling marginal $P(X_i)$: it contains $a$ itself when $X_i$ is discrete, and when $X_i$ is continuous $a$ may be taken in the interior of a region on which the label differs from $y^{*}$ . Hence $\alpha(\{X_i\},x,y^{*})>0$, which proves necessity. \hfill$\square$

\subsection{Proof of Proposition~1}

We take $Sp_k$ to be a \emph{pure} spouse of the unit $U=\{Ch_j\} \bigcup Sp(Ch_j)$: its only connection to $Y$ that is not already blocked by conditioning on $X_A$ is the collider path $Y \rightarrow Ch_j \leftarrow Sp_k$ (equivalently, $Sp_k$ enters $MB(Y)$ solely through the shared child $Ch_j$). This is required because $d$-separating $Y$ from $Sp_k$ demands that \emph{every} $Y$--$Sp_k$ path be blocked, not merely the collider path.

\textbf{(a) Before observing $Ch_j$.} Suppose $Ch_j \notin A$ and no descendant of $Ch_j$ lies in $A$. The path $Y \rightarrow Ch_j \leftarrow Sp_k$ has its collider at $Ch_j$; since neither $Ch_j$ nor any descendant of it is in the conditioning set $X_A$, the collider is closed and the path is blocked. By the pure-spouse scoping every $Y$--$Sp_k$ path is then blocked, so $X_A$ $d$-separates $Y$ from $Sp_k$. By (T1), $Y$ is conditionally independent of $X_{Sp_k}$ given $X_A$ for \emph{every} value $x_A$, so the instance-wise conditional mutual information vanishes,
\[
   I\!\left(Y;X_{Sp_k} \mid X_A=x_A\right)=0 \qquad \text{at every instance.}
\]
As $Sp_k$ then carries no information about $Y$ given $X_A$, the aligned predictor (Assumption~3) is invariant to $X_{Sp_k}$, so $\alpha(\{Sp_k\},x,y^{*})=0$. Being a blanket feature of zero necessity, $Sp_k$ lies in the semifactual set $\mathcal{S}(A)$. \hfill$\square$

\textbf{(b) After observing $Ch_j$.} Suppose $Ch_j \in A$. The collider $Ch_j$ now lies in the conditioning set, which \emph{opens} the v-structure $Y \rightarrow Ch_j \leftarrow Sp_k$. By (T2) an active path implies dependence, so the population conditional mutual information is positive, $I(Y;X_{Sp_k} \mid X_A)>0$. Unlike part~(a), this is an averaged statement: the instance-wise value $I(Y;X_{Sp_k} \mid X_A=x_A)$ is positive on a set of $x_A$ of positive probability, but need not be positive at every instance. On that set, the posterior $P(Y \mid X_A=x_A)$ varies with $x_{Sp_k}$; at an instance where this variation is decisive in the sense of Condition~(D), moving the $\argmax$ across the decision boundary. Then, perturbing $X_{Sp_k}$ flips the prediction, giving $\alpha(\{Sp_k\},x,y^{*})>0$ and placing $Sp_k$ in the counterfactual set $\mathcal{C}(A)$. Thus, once the shared child is acquired, $Sp_k$ may transition from $\mathcal{S}(A)$ to $\mathcal{C}(A)$; the driving mechanism is the explaining-away effect at the collider $Ch_j$. \hfill$\square$

\subsection{Proof of Proposition~2}

Suppose the shared child of the unit is unacquired, $Ch_j \notin A$.

\textit{The child term is positive.} $Y$ and $Ch_j$ are adjacent (the edge $Y \rightarrow Ch_j$), and adjacent vertices are not $d$-separated by any conditioning set excluding them; with $Ch_j \notin A$ the edge is an active path given $X_A$, so by (T2) $Y$ and $X_{Ch_j}$ are dependent and $I(Y;X_{Ch_j} \mid X_A=x_A)>0$ (on a positive-probability set of instances, as in Proposition~1(b)).

\textit{The spouse term is zero.} By Proposition~1(a), since $Ch_j \notin A$ and no acquired descendant of $Ch_j$ exists, $I(Y;X_{Sp_k} \mid X_A=x_A)=0$.

Combining the two,
\[
   I\!\left(Y;X_{Ch_j} \mid X_A=x_A\right)>0=I\!\left(Y;X_{Sp_k} \mid X_A=x_A\right).
\]
A policy that ranks unacquired features greedily by conditional mutual information or by mutual information per unit cost, $I(Y;X_i \mid X_A=x_A)/c(X_i)$, the spouse-selection score of Algorithm~\ref{alg:edfa}. Therefore, such policy assigns the spouse a score of exactly $0$ and the child a strictly positive score for any finite cost, and acquires the child member $Ch_j$ before any spouse of the unit. This is exactly the PC-first branch of Algorithm~\ref{alg:edfa}: the structural ordering and the information-greedy ordering coincide within a unit. Moreover the spouse's zero score is exact and holds at \emph{every} instance (part~(a)), whereas the child's positive score is only generically so; the ordering is thus a structural guarantee rather than an artefact of noisy per-instance CMI estimates, which is why the algorithm enforces PC-first directly.

By Proposition~1(b), acquiring $Ch_j$ unlocks the spouses $Sp_k$ as new counterfactual features (the $\mathcal{S} \rightarrow \mathcal{C}$ transition); acquiring the PC members first therefore increases the rate at which new recourse options (the counterfactual recourse frontier $\mathcal{R}(A)$) become available, rather than acquiring spouse members first. \hfill$\square$
\end{document}